\documentclass[10pt,journal,compsoc]{IEEEtran}

\ifCLASSOPTIONcompsoc
  \usepackage[nocompress]{cite}
\else
  \usepackage{cite}
\fi

\ifCLASSINFOpdf
\else
\fi

\usepackage[colorlinks,linkcolor=blue,anchorcolor=red,citecolor=red
,urlcolor=blue,CJKbookmarks=True]{hyperref}
\usepackage{amsmath}
\usepackage{cleveref}
\usepackage{booktabs} 
\usepackage{graphicx}
\usepackage{multirow} 
\usepackage{tabularx}
\usepackage{array}
\usepackage{amssymb}
\usepackage{color}
\usepackage{amssymb}
\usepackage{adjustbox}
\usepackage{pifont}
\usepackage{ragged2e}
\usepackage{makecell}

\usepackage{tikz}
\usetikzlibrary{backgrounds,mindmap}
\usepackage{xcolor}
\usetikzlibrary{calc,positioning,intersections}
\usepackage{pgfplots}
\usepackage{listings}

\crefname{section}{Sec.}{Secs.}
\Crefname{section}{Sec.}{Secs.}
\crefname{table}{Tab.}{Tabs.}
\Crefname{table}{Tab.}{Tabs.}
\crefname{figure}{Fig.}{Figs.}
\Crefname{figure}{Fig.}{Figs.}
\crefname{equation}{Eq.}{Eqs.}
\Crefname{equation}{Eq.}{Eqs.}

\newcolumntype{C}[1]{>{\centering\arraybackslash}p{#1}}
\begin{document}

\title{Exploring the Evolution of Physics Cognition in Video Generation: A Survey}

\author{Minghui~Lin, Xiang~Wang, Yishan~Wang, Shu~Wang, Fengqi~Dai, Pengxiang~Ding, Cunxiang~Wang, Zhengrong~Zuo, Nong~Sang, Siteng~Huang, and Donglin~Wang\\
\vspace{1ex}
\small{\textbf{Huazhong University of Science and Technology, Westlake University, Shandong University, \\
\vspace{0.2ex}
Tsinghua University, Zhejiang University}}
\IEEEcompsocitemizethanks{
\IEEEcompsocthanksitem M. Lin is with the School of Artificial Intelligence and Automation, Huazhong University of Science and Technology, Wuhan, China. Email: minghui\_lin\text@hust.edu.cn
\IEEEcompsocthanksitem X. Wang, Z. Zuo, and S. Nong are with the School of Artificial Intelligence and Automation, Huazhong University of Science and Technology, Wuhan, China. 
\IEEEcompsocthanksitem Y. Wang, F. Dai, P. Ding, D. Wang are with the School of Engineering, Westlake University, Hangzhou, China.
\IEEEcompsocthanksitem S. Wang is with the School of Control Science and Engineering, Shandong University, Jinan, China.
\IEEEcompsocthanksitem C. Wang is with Tsinghua University, Beijing, China.
\IEEEcompsocthanksitem S. Huang is with Zhejiang University, Hangzhou, China. Email: siteng.huang@gmail.com
\IEEEcompsocthanksitem Corresponding Author: Siteng Huang and Donglin Wang
}}

\markboth{Journal of \LaTeX\ Class Files,~Vol.~14, No.~8, August~2015}%
{Shell \MakeLowercase{\textit{et al.}}: Bare Advanced Demo of IEEEtran.cls for IEEE Computer Society Journals}

\IEEEtitleabstractindextext{%
\begin{abstract}
Recent advancements in video generation have witnessed significant progress, especially with the rapid advancement of diffusion models. 
Despite this, their deficiencies in physical cognition have gradually received widespread attention - generated content often violates the fundamental laws of physics, falling into the dilemma of ``visual realism but physical absurdity".
Researchers began to increasingly recognize the importance of physical fidelity in video generation and attempted to integrate heuristic physical cognition such as motion representations and physical knowledge into generative systems to simulate real-world dynamic scenarios.
Considering the lack of a systematic overview in this field, this survey aims to provide a comprehensive summary of architecture designs and their applications to fill this gap.
Specifically, we discuss and organize the evolutionary process of physical cognition in video generation from a cognitive science perspective, while proposing a three-tier taxonomy: 1) basic schema perception for generation, 2) passive cognition of physical knowledge for generation, and 3) active cognition for world simulation, encompassing state-of-the-art methods, classical paradigms, and benchmarks.
Subsequently, we emphasize the inherent key challenges in this domain and delineate potential pathways for future research, contributing to advancing the frontiers of discussion in both academia and industry. Through structured review and interdisciplinary analysis, this survey aims to provide directional guidance for developing interpretable, controllable, and physically consistent video generation paradigms, thereby propelling generative models from the stage of ``visual mimicry'' towards a new phase of ``human-like physical comprehension''.
A comprehensive list of papers studied in this survey is available at \href{https://github.com/minnie-lin/Awesome-Physics-Cognition-based-Video-Generation}{here}.

\end{abstract}

}

\maketitle

\IEEEdisplaynontitleabstractindextext

\IEEEpeerreviewmaketitle

\ifCLASSOPTIONcompsoc
\IEEEraisesectionheading{\section{Introduction}\label{sec:introduction}}
\else
\section{Introduction}
\label{sec:introduction}
\fi

\subsection{Overview}
 \IEEEPARstart{R}{ecent} years have witnessed groundbreaking advancements in video generation tasks\cite{sora,bruce2024genie,gen3,wang2024unianimate,liu2023video,yuan2023instructvideo,zhang2023i2vgen,qing2023hierarchical,wang2023videolcm,modelscopet2v,wang2024lingen,hou2024learning,lin2024stiv,chen2025goku,wang2024replace,liu2024generative,wu2024snapgen,yin2024slow,zheng2024videogen,melnik2024video,hacohen2024ltx,geng2024motion,lin2024open,polyak2024movie,lu2023vdt,wang2024recipe,chen2024livephoto,qiu2023freenoise,yang2024direct,jeong2024vmc,xing2024simda,zhao2024motiondirector,zeng2024make,xing2024dynamicrafter,jiang2025vace,zheng2024cami2v,sun2024diffusion,deng2024dragvideo,zhang2024tora,bahmani2024vd3d,zhang2024moonshot,soleimani2024survey,sun2024sora,liu2024aicl,he2023scalecrafter,he2023animate,wan2025}. These generative video models, typically trained on vast amounts of real-world video data, demonstrate remarkable capabilities in producing temporally and spatially coherent video sequences based on multimodal conditional signals (e.g., text\cite{yang2024cogvideox,singer2022make,guo2023animatediff,wu2023tune,chen2023control,ni2024ti2v}, images\cite{pan2022st,hu2022make,ren2024consisti2v,lei2024animateanything,guo2024i2v}, or videos\cite{kondratyuk2023videopoet,bar2024lumiere,wu2024fairy,wei2025dreamrelation}). 
{
These existing techniques such as Sora~\cite{sora}, Kling~\cite{kling}, and HunyuanVideo~\cite{kong2024hunyuanvideo} have demonstrated realistic visual quality, temporal continuity, and powerful capability of prompt following, and have also achieved great success in many downstream tasks, including video customization~\cite{zhao2024motiondirector,wei2024dreamvideo,jeong2024vmc,wei2025dreamrelation}, video editing~\cite{cong2024flatten,li2022physically,sun2024diffusion,hsu2024autovfx}, and video super-resolution~\cite{zhou2024upscale,jiang2024tempdiff}, etc.
} More importantly, video generation is increasingly being applied to domains such as gaming\cite{xia2024video2game,ren2025videoworld,bruce2024genie}, robotics\cite{zhao2025tasterobadvancingvideogeneration,lu2025manigaussian}, autonomous driving\cite{yang2024drivephysica,zhao2024drivedreamer4d,fu2024exploringinterplayvideogeneration}, and scientific research\cite{liu2025physgen} through techniques like instruction tuning\cite{lai2024lego}, contextual learning\cite{xing2024make}, planning\cite{huang2023diffusion}, and reinforcement learning (RL)\cite{furuta2024improving}, playing a crucial role in the development of Artificial General Intelligence (AGI). As noted by Yang et al.\cite{yang2024video}, video generation models, much like language models, are progressively evolving into autonomous agents, planners, environment simulators, and computational engines. Ultimately, video generation models have the potential to serve as artificial brains capable of reasoning and acting within the physical world.

\begin{figure*}[ht]
    \centering
    \includegraphics[width=1\linewidth]{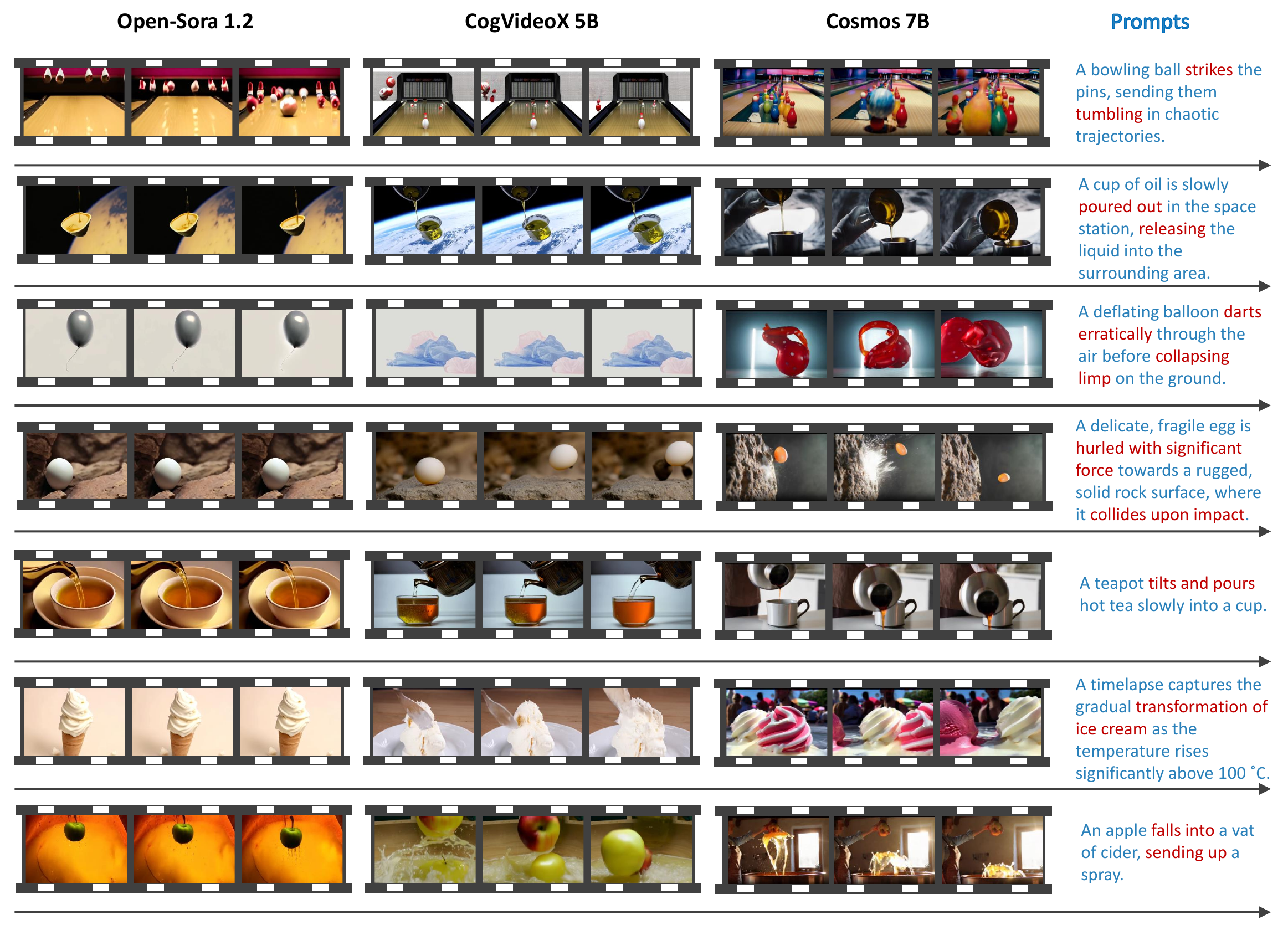}
    \vspace{-4mm}
    \caption{
    {Video cases generated by three typical state-of-the-art generative video models\cite{sora,yang2024cogvideox,agarwal2025cosmos}. We can observe that these advanced models still struggle to produce satisfying videos that strictly conform to physical laws.}
    }
    \label{fig:physics_violation}
\end{figure*}

{
Despite remarkable success,
} studies\cite{motamed2025generative,kang2024farvideogenerationworld,meng2024towards} have shown that these models often exhibit significant deficiencies in physical cognition when dealing with complex dynamic scenes. For instance, generated results frequently violate fundamental physical laws (e.g., Newtonian dynamics, momentum conservation, and energy conservation) in scenarios involving rigid body collisions, fluid dynamics, or elastic deformations, resulting in "visually realistic yet physically absurd" content, see in \cref{fig:physics_violation}. These contradictions underscore the bottlenecks in video generation models' capacity for physical cognition modeling, which may have significant negative impacts on AI applications such as robotics and autonomous driving.

{
Therefore, research on physical cognition in video generation has begun to receive widespread attention in both academia and industry~\cite{yang2024pysical,lin2024phy124,bansal2024videophyevaluatingphysicalcommonsense}. 
Recent advancements involve the systematic integration of diverse forms of physical information into generative architectures, such as motion-driven video generation and the integration of physical simulators with 3D representation-based rendering for interactive dynamics.
With the rapid progress in physical cognition in video generation, efforts to track and compare the latest research on this topic have become extremely important and meaningful.
However,
existing surveys on this topic are limited in the general AI-Generated Content (AIGC) field~\cite{meng2025grounding,cao2023comprehensive} or paying less attention on video generation~\cite{liu2025generative}.
To this end, this survey aims to fill this gap and sort out the precise and comprehensive development of physical cognition in video generation for readers.
}

To enhance the interpretability of physical cognition in video generation and strengthen its human-like capability as an artificial brain for reasoning and acting in the physical world\cite{yang2024video}, we draw inspiration from human physical cognitive mechanisms to systematically categorize physical cognition in video generation. Through this approach, we aim to provide cognition-driven solution guidance for addressing the persistent ``physical embedding bottleneck'' in video generation.

Overall, this 
{survey}
aims to establish an evolutionary pathway for modeling physical cognition in video generation from a cognitive science perspective. By providing a comprehensive and structured review of existing methods, we seek to offer guidance for developing explainable, controllable,  predictable, and physically consistent video generation paradigms.

\subsection{Taxonomy}

\begin{figure}[t]
    \centering
    \includegraphics[width=1\linewidth]{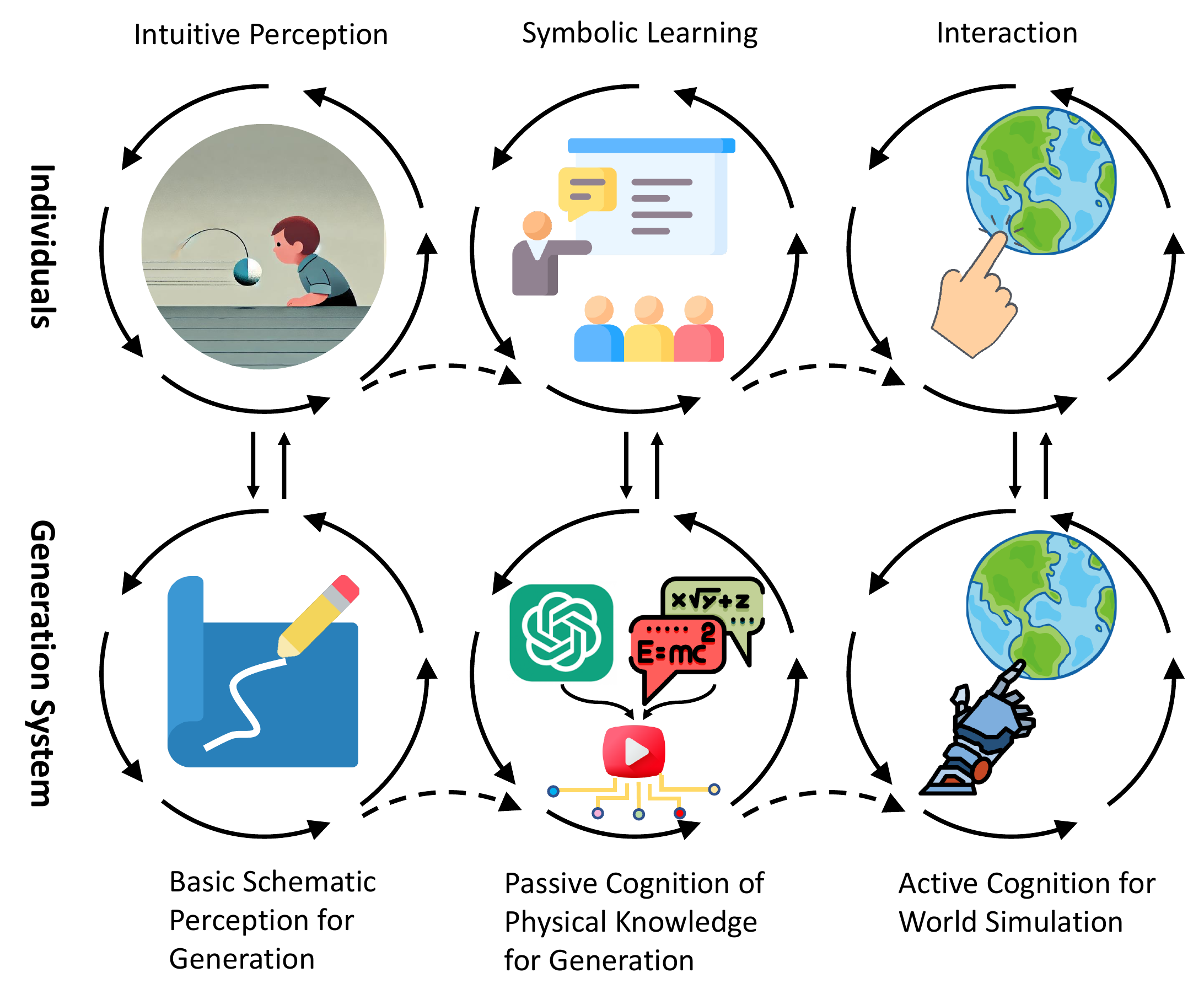}
    \vspace{-20pt}
    \caption{Cognitive evolution processes of individuals and generation system.}
    \label{fig:cognition_evolution}
\end{figure}

The evolution of cognitive systems in human development exhibits distinct stages, forming the fundamental mechanism through which individuals understand and explore the physical world. Based on Piaget's theory of cognitive development \cite{barrouillet2015theories}, we adapt and refine the characterization of an individual's cognition of the physical world, proposing that it evolves in a spiral manner through three stages: ``\textbf{Intuitive perception-Symbolic learning-Interaction}'' (\cref{fig:cognition_evolution}). In the initial stage (exemplified by infancy), individuals develop an intuitive sense of physical reality (e.g., object permanence) through primitive sensory schemas, however, this perception remains chaotic despite the omnipresence of physical principles. As cognitive development progresses to the next stage, individuals begin to acquire physical knowledge pssiavely through observation and symbolic learning (e.g., witnessing falling apples or memorizing Newtonian laws). At a more advanced stage of cognition, humans develop the ability to actively reason about and predict physical phenomena, continuously refining their cognitive models through active interaction with the environment. Contemporary video generation systems exhibit profound mapping with this evolutionary trajectory of human physics cognition. 
Video generation methods encompass different approaches, including fundamental schematic perception, passive learning of physical knowledge, and active interaction with the environment.

This {survey}
establishes an evolutionary framework for physical cognition modeling in generation systems through the lens of individual cognitive development, as shown in \cref{fig:mindmap}. We systematically categorize state-of-the-art research into three pivotal areas: \textbf{Basic Schematic Perception for Generation}: Relies on unidirectional stimulation of low-fidelity visual patterns, leading to intuitive responses (e.g., object localization without contextual awareness); \textbf{Passive Cognition of Physical Knowledge for Generation}: Grounding generation through pre-stored static physical knowledge from physics simulators or Large Language Models (LLMs); \textbf{Active Cognition for World Simulation}: Emphasizing active interaction with the environment to achieve more physically faithful future predictions. Specifically, this survey provides a systematic analysis of the following aspects:
\begin{itemize}
\item \textbf{Basic Schematic Perception for Generation (\cref{sec:schematic})}: We discuss how video and motion-based generative models integrate fundamental motion patterns to enhance motion consistency in dynamic scenes. Additionally, we explore relighting techniques and zero-shot self-guided generation approaches;
\item \textbf{Passive Cognition of Physical Knowledge for Generation (\cref{sec:passive_cognition})}: We systematically review various mechanisms for embedding physical knowledge into generative models, demonstrating how such cognitive grounding improves physical interpretability and physical consistency in generated content;
\item \textbf{Active Cognition for World Simulation (\cref{sec:active_cognition})}: We investigate generative models that predict the future through active interaction with the environment, illustrating how this approach effectively bridges the gap between video generators and real-world physical dynamics.
\end{itemize}

Finally, we discuss existing physical evaluation benchmarks and highlight unresolved challenges, such as constructing large foundational physics models, improving the physical fidelity of world simulators, incorporating multi-sensor data, enhancing the efficiency of physical simulation, addressing data scarcity and the Sim2Real gap, advancing physical quality assessment, and other related issues.

\subsection{Structure}
The {overall category structure of existing works discussed in} this {survey}
is illustrated in \cref{fig:mindmap}, and is organized as follows: In \Cref{sec:introduction}, we introduce the importance of physical fidelity in video generation and provide an overview of the classification criteria. In \Cref{sec: background}, we introduce fundamental background knowledge that encompasses physical commonsense, mainstream generative models, and physics simulators, laying the groundwork for subsequent discussions. From \Cref{sec:schematic} to \cref{sec:active_cognition}, we detail the evolutionary progression of physical cognition in video generation: In \Cref{sec:schematic}, we focuses on Basic Schematic Perception for Generation, discussing open-loop video generation methods based on fundamental representation signals such as video-based and motion-based generation; 
In \cref{sec:passive_cognition}, we explore Passive Cognition-Based Generation, emphasizing approaches that incorporate symbolic knowledge embeddings; In \Cref{sec:active_cognition}, we investigate diverse environment-interactive mechanisms, including multi-modal data-driven methods, spatial awareness, and external feedback mechanisms. In \Cref{sec:benchmark_metric}, we outline existing benchmarks and evaluation metrics used for assessing the physical plausibility of generated videos. In \Cref{sec:challenge}, we discuss current challenges and explore potential future directions in the field. Finally, in \cref{sec: conclusion}, we summarize the key findings and contributions of this survey.

\section{Survey Scope and Comparison}

\textbf{Survey Scope. }This survey focuses on the investigation of video generation methods with physical fidelity, including the generation of 2D videos, dynamic 3D, and 4D. To emphasize the physical significance in video generation, we exclude methods of unconditional video generation\cite{gupta2022rv} and long video generation\cite{li2024survey} in the general sense. Additionally, pure visual transformation methods that do not involve any individual motion, physical priors, dynamic modeling, or real-world constraints, such as art style transfer\cite{yang2022vtoonify} and video super-resolution\cite{zhou2024upscale}, are also outside the scope of this survey. Our focus is on video generation techniques that adhere to physical laws, such as kinematics, dynamics, or optical properties, to ensure visually realistic and physically plausible results. 

\textbf{Comparison with Other Surveys.} 
Unlike existing surveys on physics AI, which adopt the classification of ``explicit simulation and implicit learning''\cite{liu2025generative} or focus solely on 3D/4D generation\cite{meng2025grounding}, this survey introduces an innovative classification paradigm inspired by the human cognitive perspective. By drawing on the evolutionary trajectory of human physical cognition, we systematically reconstruct the developmental pathway of physical knowledge in video generation techniques. This classification not only provides design guidance for physics-augmented generative models but also offers multi-pathway analysis for embedding cognitive science into AGI development\cite{wang2018conceptions,voss2023we}.

\section{BACKGROUND}\label{sec: background}
This section first introduces the classification of physical commonsense in daily life and provides illustrative examples (see \cref{sec:phys_com}). Subsequently, we provide a detailed introduction to several mainstream generative models in \cref{sec:generative_models}. Finally, in \cref{sec:physics_simulation}, we explore the physics simulators, physics engines, and related platforms discussed in this survey. These aspects lay the foundation for understanding physics cognition-based generation approaches.

\begin{figure}
    \centering
    \includegraphics[width=1\linewidth]{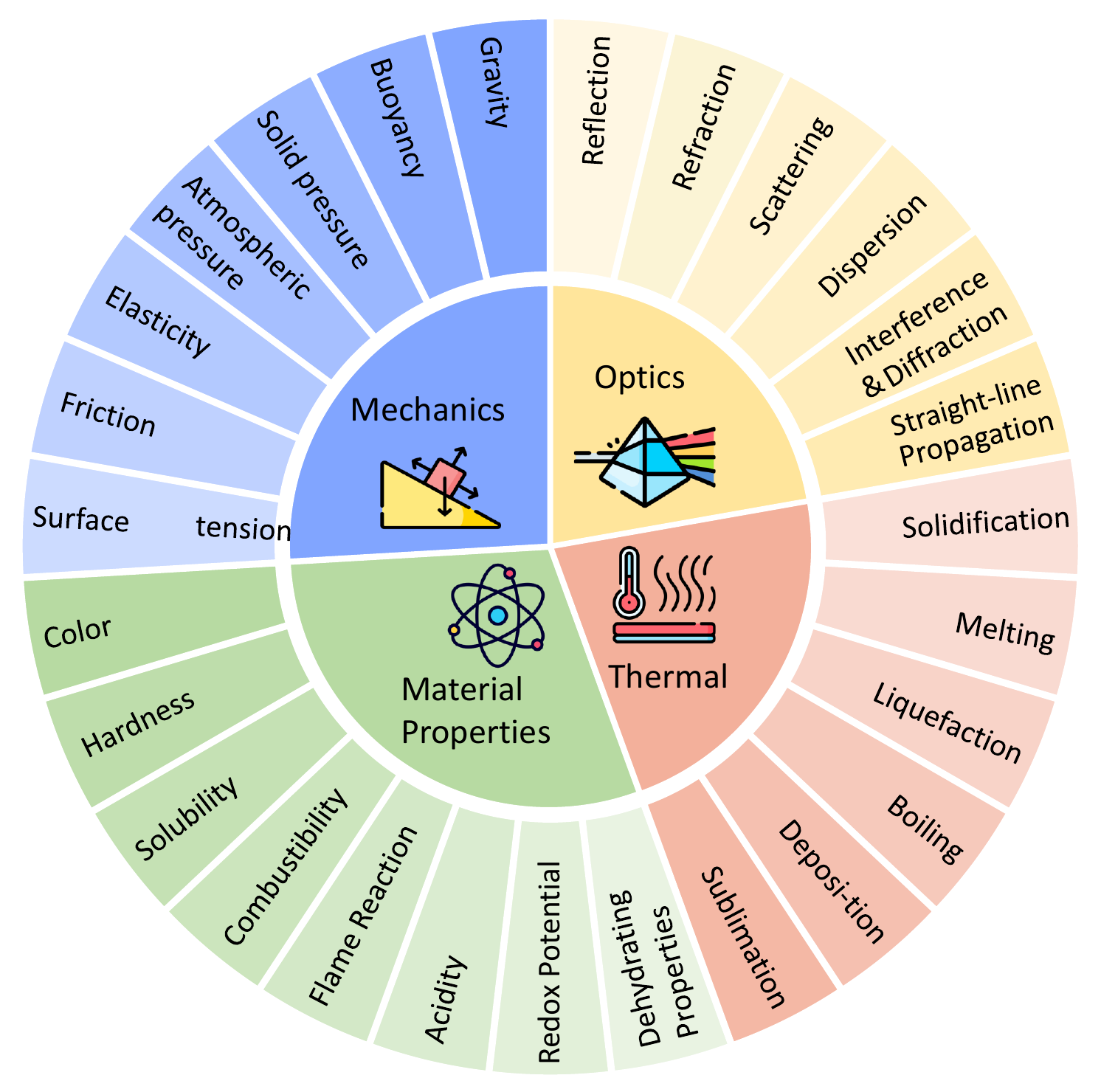}
    \caption{The taxonomy of PhysGenBench\cite{meng2024towards} benchmark, including 4 physical commonsense and 27 physical laws.}
    \label{fig:physgenbench}
    \vspace{-5pt}
\end{figure}
\subsection{Physical Commonsense}\label{sec:phys_com}
Physical commonsense refers to the basic intuitive understanding of physical objects and behaviors encountered in daily life, while physical laws are universal scientific principles used to describe consistent behaviors in nature. Physical phenomena, on the other hand, are observable events or processes caused by the interaction of physical laws\cite{meng2024towards}. The function of a physically plausible generative model is to generate physically accurate phenomena based on cues that describe physical laws. The most universal physical commonsense in the world can be divided into four main domains: mechanics, optics, thermodynamics, and material properties. For example, in \cref{fig:physgenbench}, the PhyGenBench\cite{meng2024towards} benchmark covers these four key physical domains and includes 27 physical laws (e.g., gravity and reflection).

\begin{itemize}
    \item \textbf{Mechanics:} The study of the motion of objects and the interactions of forces with objects. It includes branches such as statics, dynamics, and fluid mechanics. For example, the friction between a car's tires and the ground allows the car to start and stop.
    \item \textbf{Optics:} The study of light propagation, refraction, reflection, and interaction with objects. For example, mirrors utilize the principle of reflection to reflect light and form images.
    \item \textbf{Thermal:} The study of energy conversion, heat transfer, and temperature changes in materials. It focuses on how heat flows between objects and interacts with other forms of energy (such as mechanical or chemical energy). For example, when a pot is heated, heat is conducted through the metal to the upper part of the pot.
    \item \textbf{Material Properties:} Refers to the characteristics of a substance’s behavior and reactions under different environmental conditions. For example, salt dissolves in water, while oil does not.
\end{itemize}

\subsection{Generative Models}\label{sec:generative_models}

\begin{figure}[t]
    \centering
    \includegraphics[width=1\linewidth]{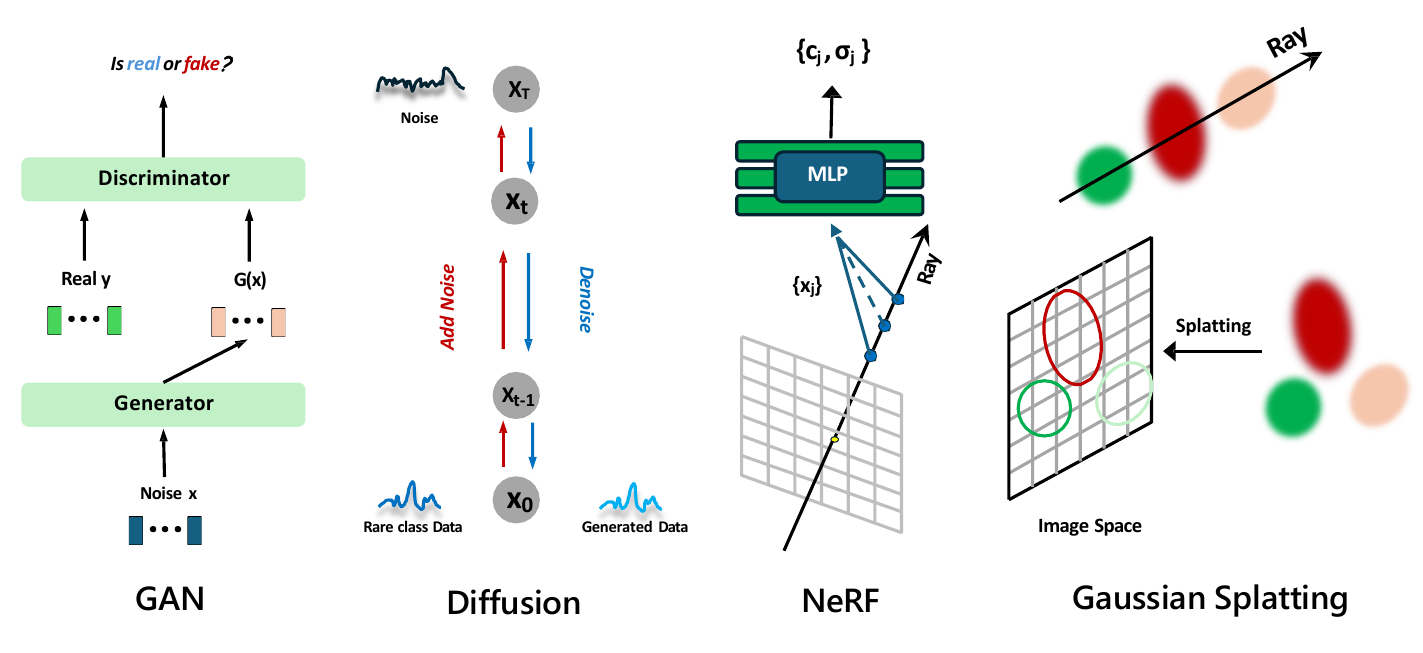}
    \caption{Introduction to mainstream generative models: GANs\cite{goodfellow2020generative}, Diffusion Models\cite{ho2020denoising}, NeRF\cite{mildenhall2021nerf}, Gaussian Splatting\cite{kerbl3Dgaussians}.}
    \label{fig:gen_models}
\end{figure}

In this subsection, we introduce several mainstream 2D and 3D generative models, {including GANs\cite{goodfellow2020generative}, Diffusion Models\cite{ho2020denoising}, NeRF\cite{mildenhall2021nerf}, Gaussian Splatting\cite{kerbl3Dgaussians}}, with a schematic illustration provided in \cref{fig:gen_models}.

\subsubsection{ Generative Adversarial Networks}
Generative Adversarial Networks (GANs)\cite{goodfellow2020generative} consist of two integral components: the Generator and the Discriminator. The Generator $G(z)$ takes random noise $z$ as input and attempts to produce samples resembling the true data distribution. Conversely, the Discriminator 
$D(x)\in [0,1]$ evaluates a given sample $x$ and outputs the probability of the sample being real. These components are trained in a minimax game manner, where the Generator aims to maximize the Discriminator's error, while the Discriminator strives to accurately differentiate between real and generated samples. The objective function, which quantifies their adversarial training process, can be expressed as:
\begin{equation}
\begin{split}
     \min_G \max_D \mathcal{L}(G, D) &= \mathbb{E}_{x \sim p_{\text{data}}(x)} [\log D(x)]\\
     & + \mathbb{E}_{z \sim p_z(z)} [\log(1 - D(G(z)))]
\end{split}
\end{equation}
This objective encapsulates the dual aims of both components, facilitating the synthesis of high-quality data through adversarial learning.

\definecolor{MossGreen}{RGB}{77,136,89}     %
\definecolor{Sand}{RGB}{237,212,170}        %
\definecolor{Clay}{RGB}{191,106,97}         %
\definecolor{Sky}{RGB}{145,191,219}         %
\definecolor{Grad3}{RGB}{255,219,185}    %
\definecolor{Grad2}{RGB}{214,234,223} 
\definecolor{Grad1}{RGB}{147,205,221}  
\definecolor{EnergeticYellow}{RGB}{255,209,0}  %
\begin{figure*}
    \centering
      \resizebox{\textwidth}{!}
    {

\begin{tikzpicture}[
    scale=0.6, %
    transform shape, %
    edge from parent fork right, grow=right, 
    level 1/.style={
        text width=6cm,
        level distance=3cm %
    },
    level 2/.style={
        text width=5.5cm,
        level distance=7cm %
    }, 
    level 3/.style={
        text width=12cm,
        level distance=6.5cm %
    }, 
    every node/.style={
        draw, 
        rectangle, 
        rounded corners = 3pt, 
        align=left,  
        inner sep=5pt, %
        font=\large %
    } %
]
    \node [rotate=90]{Physics Cognition in Video Generation}
    child {node[yshift=-6.68cm, draw=black][xshift=2cm] {Benchmarks and Metrics \ref{sec:benchmark_metric} }
        child {node[yshift=-0.05cm, draw=black] {Metrics \ref{sec:metric}}
            child{node[fill=Grad2][anchor=west][xshift=-2.7cm][minimum height=3.3em]{PhyCoPredictor\cite{chen2025phycobench}, PhyEvaler\cite{meng2024phybench},VideoCon-Physics\cite{bansal2024videocon},PhyGenEval\cite{meng2024towards}
            }}}
        child {node[yshift=0.05cm, draw=black] {Benchmarks \ref{sec:benchmarks}}
            child{node[fill=Grad2][anchor=west][xshift=-2.7cm][minimum height=3.3em]{PhyBench\cite{meng2024phybench}, VideoPhy\cite{bansal2024videophyevaluatingphysicalcommonsense}, PhyGenBench\cite{meng2024phybench}, PhyCoBench\cite{chen2025phycobench},Physion\cite{bear2021physion}, Physics-IQ\cite{motamed2025generative}, WISA-32K\cite{wang2025wisa}
            }}}
    }
    child {node[yshift=-3.95cm][xshift=2cm] {Active Cognition for World Simulation \ref{sec:active_cognition}}
        child {node[yshift=-0.1cm] {External Feedback Optimization \ref{sec:feedback}}
            child{node[fill=Grad2][anchor=west][xshift=-2.7cm][minimum height=3.3em]{
            IPO\cite{yang2025ipo}, VideoReward\cite{liu2025improving}, VideoAgent\cite{soni2024videoagent}, Gen-Drive\cite{huang2024gen},PhyT2V\cite{xue2024phyt2v}
            }}
        }
        child {node [yshift=0cm]{Spatial Environment Perception \ref{sec:spatial_env}}
            child{node[fill=Grad2][anchor=west][xshift=-2.7cm][minimum height=3.3em]{
            ManiGaussian\cite{lu2025manigaussian}, DrivePhysica\cite{yang2024drivephysica}, Abou-Chakra et al. \cite{abou-chakra2024physically}, DreMas\cite{barcellona2024dream}, DriveDreamer4D\cite{zhao2024drivedreamer4d}
            }}
        }
        child {node[yshift=0.1cm] {Multimodal Data-driven Generation \ref{sec:multimodal}}
            child{node[fill=Grad2][anchor=west][xshift=-2.7cm][minimum height=3.3em]{
            Sora\cite{sora}, Genie\cite{bruce2024genie},UniSim\cite{yang2024learning}, WorldDreamer\cite{wang2024worlddreamer}, DriveDreamer\cite{wang2023drivedreamer},GAIA-1\cite{hu2023gaia1generativeworldmodel}, Cosmos\cite{agarwal2025cosmos}
            }}
        }
    }
    child {node[yshift=1.3cm, draw=black][xshift=2cm] {Passive Cognition of Physical Knowledge for Generation \ref{sec:passive_cognition}}
        child {node[yshift=-1cm, draw=black] {LLMs Empowering Physical Simulation \ref{sec:llm_sim}}
            child{node[fill=Grad2][anchor=west][xshift=-2.7cm][minimum height=3.3em]{
            LVD\cite{lian2024llmgrounded}, AutoVFX\cite{hsu2024autovfx}, LLMPhys\cite{cherian2024llmphy}, Trans4D\cite{zeng2024trans4d},GaussianProperty\cite{xu2024gaussianproperty}
            }}
        }
        child {node[yshift=-0.8cm, draw=black] {Material Space Simulation \ref{sec:material}}
            child{node[fill=Grad2][anchor=west][xshift=-2.7cm][minimum height=3.3em]{
            DreamPhysics\cite{huang2024dreamphysics}, PAC-NeRF\cite{li2023pac}, PhysDreamer\cite{zhang2025physdreamer}, OmniPhysGS\cite{anonymous2024omniphysgs}, NeuMA\cite{caoneuma}, Feature Splatting\cite{qiu2024featuresplattinglanguagedrivenphysicsbased}
            }}
        }
        child {node[yshift=0.05cm, draw=black]{Physics Simulation-based Generation \ref{sec:physics_generate}}
            child {node[yshift=0.1cm, draw=black, text width=6cm] [anchor=west][xshift=-2.7cm] {Physicalization of Motion Field \ref{sec:phys_motion}}
                child{node[fill=Grad2, text width=7cm][xshift=-2.7cm][anchor=west][minimum height=3.3em]{PhysAnimator\cite{xie2025physanimator}, GPT4Motion\cite{lv2024gpt4motion}, PhysGen\cite{liu2025physgen}
                }}}
            child {node [yshift=0cm, draw=black, text width=6cm][anchor=west] [xshift=-2.7cm]{Interactive Dynamic Generation \ref{sec:interactive_dynamic}}
                child{node[fill=Grad2, text width=7cm][xshift=-2.7cm][anchor=west][minimum height=3.3em]{
                PhysGaussian\cite{xie2024physgaussian}, PIE-NeRF\cite{feng2024pie},VR-GS\cite{jiang2024vr}%
                }}}
        }
        child {node[yshift=0.8cm, draw=black] {Physics-Inspired Regularization \ref{sec:regularization}}
            child{node[fill=Grad2][anchor=west][xshift=-2.7cm][minimum height=3.3em]{
            GauSim\cite{shao2024gausim}, DeformGS\cite{duisterhof2024deformgssceneflowhighly}, Dynamic 3D Gaussian\cite{10550869}
            }}
        }
    }
    child {node[yshift=6.1cm, draw=black][xshift=2cm] {Basic Schematic Perception for Generation \ref{sec:schematic}}
        child {node[yshift=-0.1cm, draw=black] {Other Physical Information-guided Geration \ref{sec:other_guide}}
            child{node[fill=Grad2][anchor=west][xshift=-2.7cm][minimum height=3.3em]{
            Genlit\cite{bharadwaj2024genlit}, LumiSculpt\cite{zhang2024lumisculptconsistencylightingcontrol},
            Motion Modes\cite{pandey2024motion}, SG-I2V\cite{namekata2024sg}
            }}}
        child {node[yshift=0cm, draw=black] {Motion-guided Geration \ref{sec:motion_guide}}
            child {node[yshift=0cm, fill=Grad2] [anchor=west][xshift=-2.7cm][minimum height=3.3em]{
            Dragnuwa\cite{yin2023dragnuwa}, VideoComposer\cite{wang2023videocomposer},MotionCtrl\cite{wang2024motionctrl}, Tora\cite{zhang2024tora},  Motion-I2V\cite{shi2024motion}, Direct-a-Video\cite{yang2024direct}, C3V\cite{zhucompositional}, 3DTrajmaster\cite{fu20243dtrajmaster}
            }
            }}
        child {node[yshift=0.12cm, draw=black] {Video-guided Generation \ref{sec:video_guide}}
            child{node[fill=Grad2][anchor=west][xshift=-2.7cm][minimum height=3.3em]{
            VMC\cite{vmc}, MotionDirector\cite{zhao2024motiondirector}, SMA\cite{park2024spectral}, FlowVid\cite{liang2024flowvid}, Control-A-video\cite{chen2023control}
            }}
    }};
\end{tikzpicture}
}
\caption{Overview of the evolution of physical cognition in video generation. Please note that the typical methods listed here cover only a subset of the relevant literature and do not represent all existing studies.}
\label{fig:mindmap}
\end{figure*}
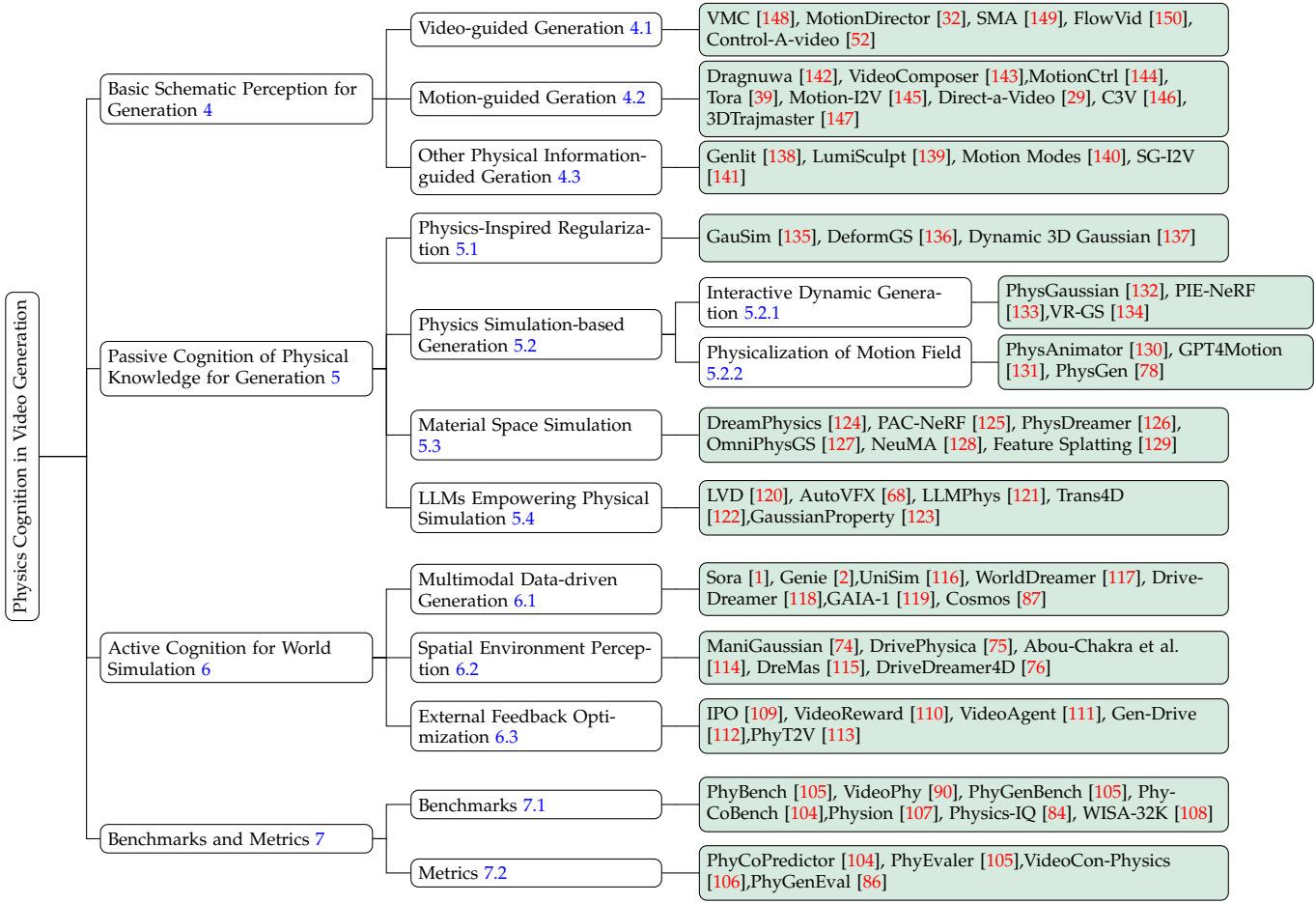

\subsubsection{Diffusion Models}
Inspired by non-equilibrium thermodynamics, diffusion model\cite{ho2020denoising} gradually adds random noise to the data distribution by defining a Markov chain of diffusion steps, and then learns the inverse denoising process to generate new data. Its core idea is to gradually destroy the data and learn to recover, contrasting with the direct adversarial training of GANs. 

The forward process (diffusion process) incrementally adds Gaussian noise to the data $ x_0$via a Markov chain until the data is transformed into pure noise $x_T$.
\begin{equation}
    q(x_t | x_{t-1}) = \mathcal{N} \left( x_t; \sqrt{1 - \beta_t} x_{t-1}, \beta_t \mathbf{I} \right)
\end{equation}
where, $\beta_t$ represents the noise level at step $t$, controlling the rate of noise addition. 

The reverse process (denoising process) trains a neural network $\epsilon_{\theta}$ to predict the noise introduced at each step, thereby generating data through iterative denoising:
\begin{equation}
    p_{\theta}(x_{t-1} | x_t) = \mathcal{N} \left( x_{t-1}; \mu_{\theta}(x_t, t), \Sigma_{\theta}(x_t, t) \right)
\end{equation}
The mean $\mu_{\theta}(x_t, t)$ and covariance $ \Sigma_{\theta}(x_t, t)$ re parameterized by a learnable neural network.

The loss function is defined as the mean squared error (MSE) for noise prediction:
\begin{equation}
    \mathcal{L} = \mathbb{E}_{t, x_0, \epsilon} \left[ \left\| \epsilon - \epsilon_{\theta}(x_t, t) \right\|^2 \right]
\end{equation}
where $\epsilon$ represents the noise added during the forward process, and $\epsilon_{\theta}$ is the neural network's prediction of this noise.
\subsubsection{Neural Radiance Fields}
Neural Radiance Fields (NeRF)\cite{mildenhall2021nerf} is a deep learning-based approach for novel view synthesis. It leverages sparse multi-view input data to train a neural network that implicitly models the volumetric density and color information of a 3D scene, enabling the synthesis of high-quality novel views. Specifically, given a spatial position \( \mathbf{x} \in \mathbb{R}^3 \) and a viewing direction \( \mathbf{d} \in \mathbb{R}^3 \) (typically represented as a unit vector), NeRF employs a continuous mapping function \( F \) to approximate the scene's volumetric density \( \sigma \) and color \( \mathbf{c} \). This function is parameterized by a neural network and is formally defined as:
\begin{equation}
F_\Theta: (\mathbf{x}, \mathbf{d}) \rightarrow (\mathbf{c}, \sigma)
\end{equation}
where $\mathbf{x} = (x, y, z) $ represents the coordinates of a point in 3D space. The viewing direction is given by $\mathbf{d} = (\theta, \phi)$, which is typically expressed in spherical coordinates. The color at the point is denoted as $\mathbf{c} = (r, g, b)$, while $\sigma$ represents the volumetric density, determining the extent to which light is absorbed at that point. The neural network parameters are denoted by $\Theta$. During the image rendering stage, NeRF employs the classical volume rendering technique, integrating color along camera rays to synthesize high-quality novel view images.

\subsubsection{Gaussian Splatting}
The core idea of Gaussian splatting\cite{kerbl3Dgaussians} is to represent a scene as a set of anisotropic 3D Gaussian kernels $G=\left\{\boldsymbol{G}_{p}:\left(\mathbf{x}_{p},\alpha_{p},\mathbf{A}_{p},\mathbf{c}_{p}\right)\right\}_{p\in\mathcal{P}}$, where each kernel is defined by its center position $\mathbf{x}_{p}$, opacity $\alpha_{p}$, covariance matrices $\mathbf{A}_{p}$, and color function $\mathbf{c}_{p}$. During the rendering phase, each Gaussian kernel is projected onto the viewpoint (i.e., splatted) and weighted accumulation is performed based on opacity and depth. The final color of the $i$-th pixel in the synthesized image from a new viewpoint is computed as follows:
\begin{equation}\mathbf{c}_{i}=\sum_{k}G_{k}(i)\alpha_{k}\mathbf{c}_{k}(\mathbf{r}_{i})\prod_{j=1}^{k-1}\left(1-G_{j}(\mathbf{i})\alpha_{j}\right).\end{equation}
where, $G_{k}(i)$ represents the projection weight of the $k$-th Gaussian kernel, and $\mathbf{r}_{i}$ is the viewpoint direction of the camera.

For 4D generation\cite{xie2024physgaussian,xia2024video2game,feng2024elastogen}, which aims to synthesize time-varying 3D scenes, simply incorporating the time variable into NeRF/3DGS\cite{mildenhall2021nerf,kerbl3Dgaussians} allows for the reconstruction and generation of dynamically evolving scenarios. Models based on NeRF/3DGS can effectively capture and synthesize realistic 4D dynamics, making them powerful tools for generating time-varying 3D scenes, such as dynamic human motions, fluid simulations, and physics-driven environments.

\subsection{Physics Simulation}\label{sec:physics_simulation}
In physics modeling, physics simulators serve as one of the critical components, translating physical laws into computable forms through numerical methods to enable precise simulation of object and environmental dynamics in complex scenarios. Meanwhile, to meet diverse application demands, simulation engines and platforms integrating multiple physics simulators have emerged. These engines and platforms unify various physics simulation functionalities through standardized interfaces and modular designs, offering more flexible and efficient solutions. In the following sections, we provide a detailed discussion of the mainstream physics simulation methods (see \cref{sec:physics_sim}), along with the leading physics engines and platforms (see \cref{sec:phys_engine}).

\subsubsection{Physics Simulation Methods}\label{sec:physics_sim}
\begin{table*}[ht]
\scriptsize
\centering
\renewcommand{\arraystretch}{1.1}
\resizebox{\textwidth}{!}{%
    \begin{tabular}{m{2cm}<{\centering}|m{2cm}<{\centering}m{1.7cm}<{\centering}m{4cm}<{\centering}m{0.8cm}<{\centering}m{4cm}<{\centering}} \toprule
         Physics Engines and Platforms&  Programming Language&  GPU Acceleration&  Supported Physics Types&  Open Source& Typical Application Scenarios\\ \midrule 
         Bullet\cite{coumans2015bullet}&  C++&  Partial support&   Rigid, soft body , collision detection, constraint solving&  $\checkmark$& Game development, film special effects, and robotics simulation\\ \midrule  
         PyBullet\cite{pybullet}& Python &  Partial support & Rigid, soft body , collision detection, constraint solving & $\checkmark$ &Robot simulation, reinforcement learning \\ \midrule 
         Blender Physics\cite{lv2024gpt4motion}& Python &  Partial support& Rigid body, soft body, fluids, fabrics and particle systems & $\checkmark$ &3D animation and film special effects \\ \midrule  
         Isaac Gym\cite{makoviychuk2021isaac} & Python & Fullly support & Rigid body and joint drive & $\times$ & Large-scale parallel simulation, complex robot motion control\\ \midrule
         NVIDIA PhysX\cite{physx} & C++ & Fullly support & Rigid body, soft body, cloth, fluids and particle system & $\times$ & Game development, virtual reality, and industrial simulation\\ \midrule
         Taichi\cite{hu2020difftaichi}& Python/C++ & Fullly support & Fluids, elastic bodies, particle systems & $\checkmark$ &Support for custom physical models (e.g. MPM\cite{jiang2016material}) \\ \midrule 
         NVIDIA Omniverse\cite{omniverse}& Python/C++/USD\cite{usd} &Fullly support  & Multi-physics engine integration& $\times$ & Cross-software collaboration, digital twins \\ \midrule 
         Genesis\cite{Genesis}& Python & Fullly support & Coupling various physical models & $\checkmark$ &Robot realistic data generation \\ \bottomrule
    \end{tabular}
    }
    \vspace{3pt}
    \caption{Comparison of features of physics engines and platforms.}
    \label{tab:engines}
    \vspace{-20pt}
\end{table*}
Currently, mainstream physical simulation approaches include grid-based Lagrangian-Eulerian hybrid methods (e.g., the Material Point Method, MPM\cite{jiang2016material,xie2024physgaussian}, widely used for simulating elastomers, fluids, and granular materials) and positon-based methods (e.g., Position-Based Dynamics, PBD\cite{muller2007position},XPBD\cite{macklin2016xpbd}, commonly applied in real-time simulations of deformable objects and cloth).

\textbf{Position-Based Dynamics, PBD}\cite{muller2007position}:
PBD\cite{muller2007position} adjusts particle positions using constraint conditions defined as cost functions, which are optimized to satisfy the desired constraints. In a typical dynamic system, each particle (or vertex) is assigned a mass, position, and velocity. The simulation process involves predicting new positions based on current velocities and external forces, iteratively correcting these positions to enforce the constraints, and then updating velocities accordingly.

\textbf{Extended Position-Based Dynamics, XPBD}\cite{macklin2016xpbd}: XPBD is an extension of the PBD\cite{muller2007position} framework, designed to address several inherent limitations of PBD, such as time step dependency and imprecise control over constraint stiffness. The core idea of XPBD is to model constraints as elastic potential energy and solve constraint forces using implicit integration methods. This approach maintains the computational efficiency of PBD while delivering more accurate and physically plausible simulations.

\textbf{Material Point Method, MPM}\cite{jiang2016material,xie2024physgaussian}: The MPM combines the Eulerian and Lagrangian approaches to solve PDEs by discretizing space into a grid-particle framework, thereby simulating the motion and deformation of materials at macroscopic scales. Specifically, in MPM, an object is represented as a collection of discrete particles, each carrying its own material properties (e.g., mass, density, and velocity). The simulation alternates between transferring particle data to a fixed grid (P2G) for computing forces and updating grid-based quantities, and then transferring the updated information back to the particles (G2P) to revise their positions and velocities. This iterative process allows MPM to accurately simulate complex deformations and interactions in a physically consistent manner.

Notably, compared to PBD\cite{muller2007position} and XPBD\cite{macklin2016xpbd}, MPM\cite{jiang2016material} is a differentiable simulator. Its mathematical differentiability enables a deep integration between physical systems and data-driven methods (e.g., neural networks). Such frameworks can generate high-quality videos that strictly adhere to physical laws, offering both visual fidelity and dynamic realism.

\subsubsection{Physics Engines and Platforms}\label{sec:phys_engine}

In this subsection, we will introduce several mainstream physics engines and platforms that are covered in this survey, as shown in \cref{tab:engines}.

\textbf{Bullet Physics}: Bullet\cite{coumans2015bullet} is a robust and open-source physics engine widely utilized in game development, film special effects, and robotics simulation. It supports rigid body dynamics, soft body dynamics, collision detection, and constraint solving, boasting high performance and cross-platform compatibility. 

\textbf{PyBullet}: PyBullet\cite{pybullet} is a Python interface encapsulation of the Bullet\cite{coumans2015bullet}, designed to offer a user-friendly programming experience. It not only supports complex physical simulations but also integrates graphic rendering functionalities, enabling the easy creation of robotic simulation environments and reinforcement learning scenarios.

\textbf{Blender}: Blender\cite{lv2024gpt4motion} is a comprehensive open-source 3D creation software that integrates modeling, animation, rendering, and video editing features. Its built-in physics engine supports simulations of rigid bodies, soft bodies, fluids, fabrics, and particle systems. Although it may not match the performance of specialized physics engines, it excels in artistic creation and visual effect production.

\textbf{Isaac Gym}: Isaac Gym\cite{makoviychuk2021isaac} is a high-performance physics simulation platform developed by NVIDIA, specifically designed for reinforcement learning and robotics simulation. Utilizing GPU acceleration technology, it can run thousands of simulation environments in parallel on a single GPU, significantly enhancing training efficiency. 

\textbf{NVIDIA PhysX}: NVIDIA PhysX\cite{physx} has emerged as a predominant solution in gaming and real-time simulation domains by achieving real-time computation of large-scale physical interactions through GPU optimization, effectively balancing computational accuracy with system performance.

\textbf{Taichi}: Taichi\cite{hu2020difftaichi} is an open-source library for computer graphics and physical simulations, focusing on high-performance computing and programmability. It employs Just-In-Time (JIT) compilation technology to convert Python code into highly efficient low-level code, supporting both CPU and GPU operations. 

\textbf{NVIDIA Omniverse}: NVIDIA Omniverse\cite{omniverse} is a real-time 3D design collaboration and physics simulation platform built on the Universal Scene Description (USD)\cite{usd} framework. Its core features include cross-software collaboration, physics-level precision simulation (digital twins), and a generative AI-driven toolchain, aimed at advancing the development of generative physical AI.

\textbf{Genesis}: Genesis\cite{Genesis} is an open-source generative physics engine designed for robotics, embodied intelligence, and physical AI. Built in Python, it integrates multiple physics solvers\cite{macklin2016xpbd,jiang2016material,muller2007position} into a unified framework, enabling fast simulations. A key feature is its differentiable simulators, which support reinforcement learning by embedding gradient information. Genesis also supports language-driven generative simulations, allowing the creation of high-fidelity 4D dynamic worlds.

\section{Basic Schematic Perception for Generation}\label{sec:schematic}

Although traditional video generation models demonstrate remarkable generative capabilities, their mechanisms for physical dynamic control remain significantly constrained. These models typically rely on simple signal conditioning (e.g., text\cite{yang2024cogvideox,singer2022make,guo2023animatediff,wu2023tune} or images\cite{pan2022st,hu2022make,ren2024consisti2v}), which inadequately specifies fine-grained dynamic details. To address this limitation, recent approaches abstract physical visual patterns into controllable basic schemas, such as optical flow fields\cite{liang2024flowvid}, trajectories\cite{yin2023dragnuwa,wang2024motionctrl}, motion bounding boxes\cite{wei2024dreamvideo}, or motion video sequences\cite{zhao2024motiondirector,vmc}. These basic schemas encode the temporal evolution of object entities or pixel positions as motion fields, which are then injected into the model's latent space, ultimately achieving motion-consistent video generation. This section's paradigm pipeline is shown in \cref{fig:motion_signal_guide}.

\subsection{Video-guided Generation}\label{sec:video_guide}

\begin{figure}
    \centering
    \includegraphics[width=0.9\linewidth]{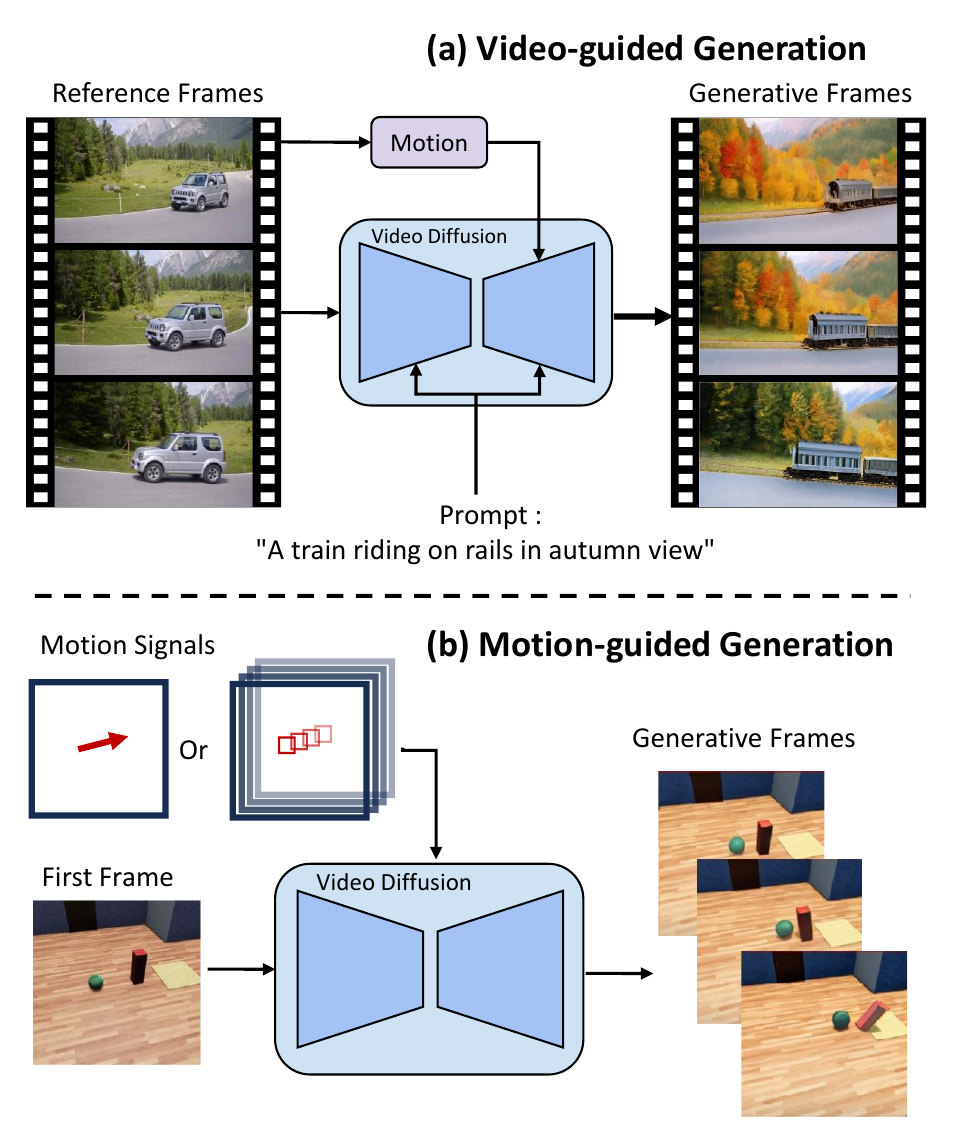}
    \caption{Basic motion signal-guided generation pipeline. (a) is a video-guided generation pipeline in \cref{sec:video_guide}, while (b) is a motion-guided generation pipeline in \cref{sec:motion_guide}.}
    \label{fig:motion_signal_guide}
\end{figure}

The core objective of Video-to-Video (V2V) generation lies in extracting and transferring motion attributes (e.g., motion direction, velocity distribution, and temporal pose evolution) from reference videos while reconstructing the visual environment of target scenes. As shown in \cref{fig:motion_signal_guide}(a), current methodologies achieve this through a dual-prior driven framework: reference videos provide motion pattern priors (such as rigid-body motion continuity and temporal coherence of human actions), while text/image prompts supply content semantic priors (e.g., scene layouts, object materials). 

\textbf{Unwrapping Motion and Appearance.} To address the complex entanglement between motion and appearance, some studies focus on separating and refining motion features. For example, VMC (Video Motion Customization)\cite{vmc} aligns the ground truth residuals between consecutive frames with their denoising predictions, enabling the distillation of motion information and fine-tuning the keyframe generation module to improve temporal consistency. At a higher level, MotionDirector\cite{zhao2024motiondirector} adopts a dual-path architecture to decouple appearance and motion features from reference videos. This method not only reproduces the same motion across different appearances but also supports the blending of motion and appearance features from different videos, enhancing the diversity and controllability of generation. Spectral Motion Alignment (SMA)\cite{park2024spectral} addresses the issues of contextual information deficiency and local distortions that may arise from motion vectors obtained through traditional inter-frame residuals by coordinating global and local frequency domain regularization. Sun et al.\cite{sun2024video} disentangles appearance and motion by distinguishing frame-level and spatiotemporal representations, achieving motion control by minimizing appearance information.

\textbf{Extracting Motion Signals.} Additionally, some studies the concept of explicit motion correlation between reference and generated videos. These approaches typically extract conditional signals from the reference video and embed them into the generation framework for learning. For instance, FlowVid\cite{liang2024flowvid} addresses the limitations of conventional optical flow constraints by introducing depth maps as spatial conditions to assist motion control, significantly improving spatial consistency. Control-A-Video\cite{chen2023control} combines motion priors from the reference video’s motion flow and inter-frame residuals to warp latent noise, thereby controlling motion consistency across generated video frames. 
Li et al.\cite{li2024_GenerativeImageDynamics} focuses on natural motion by extracting motion spectral volume\cite{davis2016visual} from real-world videos and modeling them as oscillatory patterns. These patterns are used to guide the generation of seamless looping videos or interactive dynamic simulations. Wang et al.\cite{wang2024motion} integrates the strengths of explicit and implicit methods by abstracting the motion information from reference videos into Q, K, and V embeddings within temporal attention. It uses an optical flow-like inter-frame difference approach to eliminate static appearance biases, capturing global and local motion patterns across the temporal dimension and further ensuring both temporal and motion consistency.

\textbf{Causal Reasoning Outside of Motion.} In scenarios involving complex interactive dynamics, InterDyn\cite{akkerman2024interdyn} introduces hand control signals in the form of masks, which are processed through a ControlNet-like\cite{zhang2023adding} branch network to drive action generation. This approach enables the prediction of complex interactions involving counterfactual future scenarios and force propagation dynamics. Unlike traditional methods, InterDyn leverages the implicit physical priors in pre-trained video generation models, allowing for continuous causal dynamic reasoning beyond the scope of the control signals. Although it does not require explicit reconstruction or physical simulation, it still relies on pre-defined motion signals to guide subsequent predictions.

\subsection{Motion-guided Generation}\label{sec:motion_guide}

Video-to-video (V2V) generation methods are primarily limited to reproducing existing motion patterns. To achieve greater flexibility in adapting to the generation of diverse motion patterns, the research community has increasingly focused on motion-guided generation paradigms-introducing explicit motion control signals as conditional constraints to steer models toward generating controllable dynamic content. 
Motion-guided video generation (see \cref{fig:motion_signal_guide}(b)) typically adopts either a single-stage\cite{wei2024dreamvideo,wang2024motionctrl,geng2024motion,yang2024direct} or two-stage paradigm\cite{zhang2024tora,wu2025draganything,zhou2024trackgo,yin2023dragnuwa,niu2025mofa,shi2024motion,lei2024animateanything,li2024image}. The two-stage paradigm involves: 1) extracting low-dimensional motion representations (e.g., optical flow fields, motion vectors, or point trajectories) from mask bounding boxes or arrow-guided trajectories; and 2) augmenting pre-trained video diffusion models with motion encoders to embed these motion features into the latent space of the diffusion process. In contrast, the single-stage framework omits the explicit motion extraction step, instead directly mapping simplified motion control signals to the generative model through end-to-end learning (equivalent to Stage 2 in two-stage frameworks).
To enable diffusion models to effectively respond to controllable signals, various approaches have been proposed. Some integrate conditional signals and latent representations through attention mechanisms\cite{zhang2024tora,zhou2024trackgo,lei2024animateanything,yang2024direct,shi2024motion}, while others\cite{wei2024dreamvideo,wang2024motionctrl,wu2025draganything,yin2023dragnuwa,niu2025mofa,geng2024motion,li2024image} employ additional structures similar to ControlNet\cite{zhang2023adding}, encoding external conditions into the latent space via multi-scale skip connections.  

Despite progress in generating controllable motion-driven videos under these paradigms, challenges remain, including motion inconsistencies, ambiguities in camera transitions and object movements, and difficulties in modeling 3D scene dynamics. In the following sections, we introduce improvements to motion-guided video generation across three critical dimensions: motion consistency enhancement, camera-object motion control, and 3D spatial motion generation.

\textbf{Motion Consistency Enhancement.} Motion consistency directly impacts the physical plausibility of generated videos. Ensuring motion consistency allows an object's velocity, direction, and acceleration to remain coherent across multiple frames, avoiding interruptions or unrealistic changes in motion. This enhances the video's visual coherence and viewing experience. Current research focuses on designing motion signals to drive the generation of physically plausible motion videos.

Initially, Dragnuwa\cite{yin2023dragnuwa} achieved controllable video generation across semantic, spatial, and temporal dimensions by integrating three fundamental control signals: text, images, and trajectories. Through an end-to-end pipeline, it sampled dense optical flow directly from video streams and fused it with text and image features at multiple scales, adaptively training to generate trajectory-consistent videos. 
{
VideoComposer~\cite{wang2023videocomposer} presented a compositional video generation framework and leveraged motion vectors as a temporal condition for motion transfer.
Animate Anyone~\cite{hu2024animate} and UniAnimate~\cite{wang2024unianimate} took a character image as a reference and used a pose sequence to control the desired movement of the character.
}
MotionCtrl\cite{wang2024motionctrl} introduced independent modules for controlling camera and object movements, enabling 2D point-driven object motion and 3D trajectory-driven camera control. MOFA-Video\cite{niu2025mofa} and Motion Dreamer\cite{xu2024motiondreamer} used a sparse-to-dense motion generation network to extract explicit motion cues and incorporated various motion fields as control conditions to fine-tune the diffusion model. Motion prompting\cite{geng2024motion} expands user requests into point trajectory motion prompts, encoding both sparse and dense motion in spatial and temporal dimensions. Motion-I2V\cite{shi2024motion} employed explicit motion modeling to generate videos of moving objects while preserving their appearance. The method introduced a diffusion-based motion field predictor and a motion-guided I2V generator, leveraging a temporal attention module to connect the two stages and propagate motion trajectories consistently across synthesized frames. Dreamvideo-2\cite{wei2024dreamvideo} introduced reference attention to learn subject appearance features, while bounding box mask sequences provided motion control signals. To achieve a balance between subject representation and motion control, the method masked reference attention to enhance subject identity representation. Tora\cite{zhang2024tora} built upon Sora\cite{sora} and further emphasized the importance of motion control by incorporating dynamic trajectory components to guide scalable video generation. 

Earlier trajectory-controlled motion methods relied solely on single-point controls to generate pixel-level motion, which struggled to produce realistic physical entity movements. These limitations often resulted in undesired global motion or appearance distortions. To address these challenges, DragAnything\cite{wu2025draganything} defined sequences of entity-centric points and Gaussian maps to guide motion at the entity level. TrackGo\cite{zhou2024trackgo} combined masks and trajectory tracking to obtain detailed multi-point trajectories and introduced a TrackAdapter to manipulate dual-branch attention maps for motion intervention. Additionally, the application of spatiotemporal attention embedding may be misled by irrelevant blocks. Flatten\cite{cong2024flatten} utilizes optical flow to connect patches across different frames, guiding accurate attention and promoting information exchange between patches along the same trajectory.

When multiple trajectories in dynamic motion come into close proximity (e.g., collisions or overlaps), sparse optical flow may merge and erroneously swap trajectories, as observed in Tora\cite{zhang2024tora}. To mitigate this issue, InTraGen\cite{liu2024intragen} introduced multi-modal interaction encoding to ensure stable generation of multi-object interaction videos. The method designed a target ID map to provide interactive priors for sparse dynamic information, enabling the model to accurately identify both static and dynamic objects under trajectory-controlled conditions.

\textbf{Camera-object Motion Control.} Given the limitations of earlier methods in handling only a single type of motion signal, AnimateAnything\cite{lei2024animateanything} integrates multiple conditions (e.g., motion annotations, camera motion) to control video generation. By combining explicit and implicit signal injection, this method leverages a Flow Generation Model (FGM) to iteratively denoise and transform various dynamic signals into a unified dense optical flow, which is ultimately used to guide video generation. While both Motion-I2V\cite{shi2024motion} and AnimateAnything adopt a two-stage architecture to incorporate control signals, AnimateAnything stands out by enabling the integration of multiple conditions and providing balanced guidance for video generation. Similarly, Image Conductor\cite{li2024image} focuses on generating camera transitions and object motion but takes a different approach than AnimateAnything. Specifically, it decouples camera and object motion through distinct LoRA weights, allowing precise control over either motion type. On the other hand, Direct-a-Video\cite{yang2024direct} employs a self-supervised training strategy to flexibly decouple camera and object motion without pretraining. This is achieved by cropping static-camera video sequences to simulate moving-camera perspectives and embedding them into temporal cross-attention. Additionally, it emphasizes the spatial attention of text tokens corresponding to motion bounding boxes, enabling independent or joint control over camera and object motion.

\textbf{3D Spatial Motion Generation.} 
The motion trajectories of objects in the real physical world are inherently three-dimensional dynamic processes, involving complex spatial interactions such as depth displacement, multi-view occlusion, rotational and orbital motions. However, existing motion control methods are predominantly confined to two-dimensional planar trajectory guidance, struggling to accurately characterize physical dynamics scene within 3D spatial. To obtain well-defined 3D trajectories, LeviTor\cite{wang2024levitor} introduces a novel approach that combines depth information with K-means clustering of multiple points to control 3D object trajectories in video synthesis, thereby guiding the model to generate physically plausible videos in 3D space. Chen et al.\cite{chen2025physicalunderstandingvideogeneration} extends 2D conditional optical flow to a 3D point cloud motion field, achieving cross-dimensional feature fusion by concatenating video frames and 3D point vectors and feeding them into a diffusion model for training, thereby enhancing the dynamics of video generation at a higher level. C3V\cite{zhucompositional} relies on textual priors processed through LLMs to obtain object trajectories in 3D space, optimizing spatial layout and temporal dynamics in video generation. This method decomposes the input text into sub-prompts, which are processed by expert models to generate 3D representations. It then incrementally estimates object trajectory coordinates in the scene and refines dynamic objects using Score Distillation Sampling (SDS)\cite{poole2022dreamfusion} to extract 2D diffusion priors, achieving coarse-to-fine text-driven 3D video generation. TC4D\cite{bahmani2025tc4d} decomposes the deformation field into global and local components, modeling global motion as rigid body transformations along trajectories. It introduces video score distillation sampling to activate the internal motion priors of pretrained generative models, optimizing the local deformation field. This approach achieves significant progress in generating large-scale scenes.

Previous approaches either focus on obtaining 3D trajectories or directly guide video generation using trajectories in 3D space, while overlooking the issue of the lack of realistic 3D trajectory data during training. 3DTrajMaster\cite{fu20243dtrajmaster} addresses this gap by introducing the first 360°-Motion Dataset specifically designed for 3D trajectory-guided video generation. Furthermore, 3DTrajMaster develops a plug-and-play 3D trajectory embedding framework, which incorporates domain adapters and an annealed sampling strategy to enable controllable video generation with multi-entity 6DoF (six degrees of freedom).

\subsection{Other Physical Information-guided Generation}\label{sec:other_guide}
\textbf{Video Relighting}. In addition to motion realism, lighting consistency is another indispensable aspect of physical plausibility in high-fidelity video generation. Previous models for lighting modeling primarily focused on relighting for single images, either through explicit inverse rendering\cite{li2022physically,wang2021learning} of geometry and materials or by employing neural methods\cite{kocsis2024lightit,zeng2024dilightnet,ren2024relightful} for end-to-end learning. GenLit\cite{bharadwaj2024genlit} uniquely explores the potential of video diffusion models to comprehend the physical world, particularly lighting. It reformulates the single-image relighting task as a video generation task, enabling controllable light manipulation in video generation through external light source signals. LumiSculpt\cite{zhang2024lumisculptconsistencylightingcontrol} shifts the focus of the relighting task to text-to-video generation, decoupling lighting representation from appearance and enabling the diverse generation of videos with controllable lighting. Lux Post Facto\cite{mei2025lux} encodes High Dynamic Range (HDR) maps into ``lighting embeddings'' for refined lighting control, integrating them as conditional inputs into a diffusion model. By training delighting and relighting models on a hybrid dataset with lighting-rich and motion-rich, this approach achieves more precise and nuanced video relighting.

\textbf{Zero-Shot Self-guidance}. Previous methods\cite{yin2023dragnuwa,chen2023control,fu20243dtrajmaster} that leverage the generative capabilities of pre-trained models often rely heavily on conditional guidance to produce controllable motion videos, overlooking the latent motion priors embedded within pre-trained models. Motion Modes\cite{pandey2024motion} aims to uncover the motion generation potential of pre-trained diffusion models by guiding energy-based iterative sampling during inference to construct a motion set, achieving diverse motion generation without additional training or conditional control. Similarly, SG-I2V\cite{namekata2024sg} optimizes noise latent variables to enforce feature similarity within bounding box trajectories, enabling self-guided trajectory control.

\textbf{Discussion.} Overall, basic schematic perception-based generation methods highlight the effectiveness of control signals in controllable video generation tasks. By embedding various forms of schematic representations, these approaches address key challenges such as motion consistency, camera-object control, 3D spatial motion, and relighting. However, they face several critical limitations: 1) \textbf{Lack of Physical Consistency}: Generation methods guided solely by simple visual patterns often fail to capture fundamental physical principles, resulting in physically implausible video outputs; 2) \textbf{Limited Generalization to Complex Scenes}: These models struggle to adapt to diverse environments involving complex object interactions, such as collisions, deformations, and external forces. To enhance physical plausibility and improve video realism, future research could explore the integration of physical knowledge into generative frameworks or leverage real-world physical interactions to enrich the modeling of dynamic environments.

\section{Passive Cognition of Physical Knowledge for Generation}\label{sec:passive_cognition}
In addition to the aforementioned schema-guided intuitive video generation methods, the generative system can also leverage physical rules and strictly follow symbolic logic for deductive reasoning. The sources of symbolic knowledge include physical loss function constraints(in \cref{sec:regularization}), knowledge embedding from physics simulators(in \cref{sec:physics_generate}), and world physics knowledge from LLMs (in \cref{sec:llm_sim}). Explicit physics knowledge embedding methods demonstrate unique advantages in complex interactive scenarios: by directly embedding physical knowledge, the generated videos not only ensure visual coherence but also achieve physical verifiability at both the microscopic particle motion and macroscopic energy transfer levels, thus laying a theoretical foundation for building reliable world simulators.

\subsection{Physics-Inspired Regularization}\label{sec:regularization}
\begin{figure}
    \centering
    \includegraphics[width=0.8\linewidth]{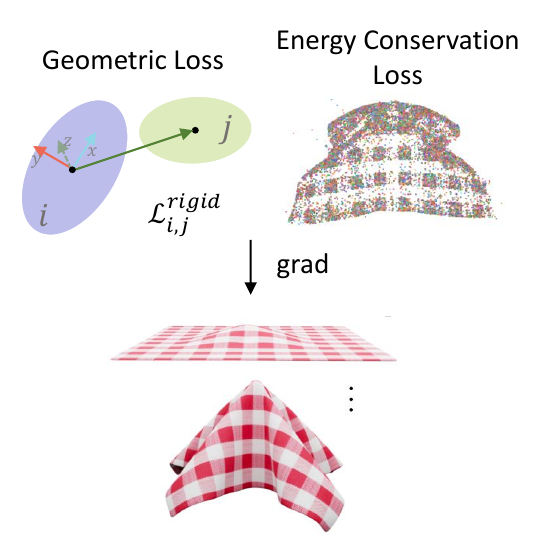}
    \vspace{-10pt}
    \caption{Physics-inspired regularization pipeline.}
    \label{fig:regular}
\end{figure}

Physics-informed regularization adds constraint terms related to physical systems within the loss function, enabling the model to learn data features while adhering to known physical laws, thereby enhancing its generalization ability and physical consistency, as shown in \cref{fig:regular}.

\textbf{Regularization of Energy Conservation.} To enable the model to more accurately reflect long-term complex deformations in the real world, GauSim\cite{shao2024gausim} combines continuum mechanics with neural networks, effectively capturing dynamic laws through hierarchical simulations from coarse to fine. Additionally, explicit constraints for mass conservation and momentum conservation are introduced to ensure physical interpretability. DeformGS\cite{duisterhof2024deformgssceneflowhighly} focuses on large deformations, shadows, and occlusion in highly deformable objects, shaping the 3DGS deformation field by optimizing deformation functions, while introducing local isometric loss and momentum conservation regularization terms to constrain the relative distance and trajectory smoothness of adjacent Gaussians. 

\textbf{Regularization of Geometric.} In dynamic scene modelingtasks, Dynamic 3D Gaussians\cite{10550869} relies on regularization to enforce underlying physical models, integrating explicit modeling with a physical metric space. Dynamic 3D Gaussians\cite{10550869} are designed for complex motion scenarios, incorporating local-rigidity loss (to enforce rigid body transformations), local-rotational similarity loss (to maintain local rotational consistency), and long-term isometric loss to achieve dense 6-DOF tracking. Dynamic 3D Gaussians and DeformGS provide high-precision 3D tracking solutions for deformable object manipulation and complex motion scenarios. Rai et al.\cite{rai2024enhancing} significantly enhances the temporal consistency and rigidity of shape preservation in animations by incorporating Length-Area (LA) regularization and shape-preserving As-Rigid-As-Possible (ARAP) loss. While incorporating 3D point clouds into the diffusion framework, Chen et al.\cite{chen2025physicalunderstandingvideogeneration} jointly optimizes the noise distribution and local rigidity preservation of the point cloud structure, thereby enhancing the geometric fidelity and physical coherence of generated content. 

 While these methods generally demonstrate good performance under predefined settings, they cannot guarantee an accurate capture of unlearned physical phenomena, such as complex interactions, which may lead to physically implausible results. Therefore, further exploration is needed for more explicit physical video generation.

\subsection{Physics Simulation-based Generation}\label{sec:physics_generate}

Physics simulation-based video generation leverages explicit physical rules and models, embedding domain-specific knowledge (e.g., dynamics, fluid mechanics, and material science) to simulate motion and interactions within a scene. In \Cref{sec:interactive_dynamic}, we investigate interactive dynamic generation methods that employ external forces as inputs for static 3D scenes. These methods model the interactions between scenes and forces using various physical simulators, infer changing trends that comply with the physics laws, and generate corresponding dynamic scenes. In \cref{sec:phys_motion}, we introduce approaches for the physicalization of the motion field. In \Cref{sec:motion_guide}, we have discussed motion-guided generation based on manually defined motion signals, which inherently exhibit high uncertainty. To address this, the methods described in \cref{sec:phys_motion} leverage physics simulators to generate precise motion fields, effectively guiding the diffusion model for more physically accurate video synthesis.

\begin{figure}[t]
    \centering
    \includegraphics[width=1\linewidth]{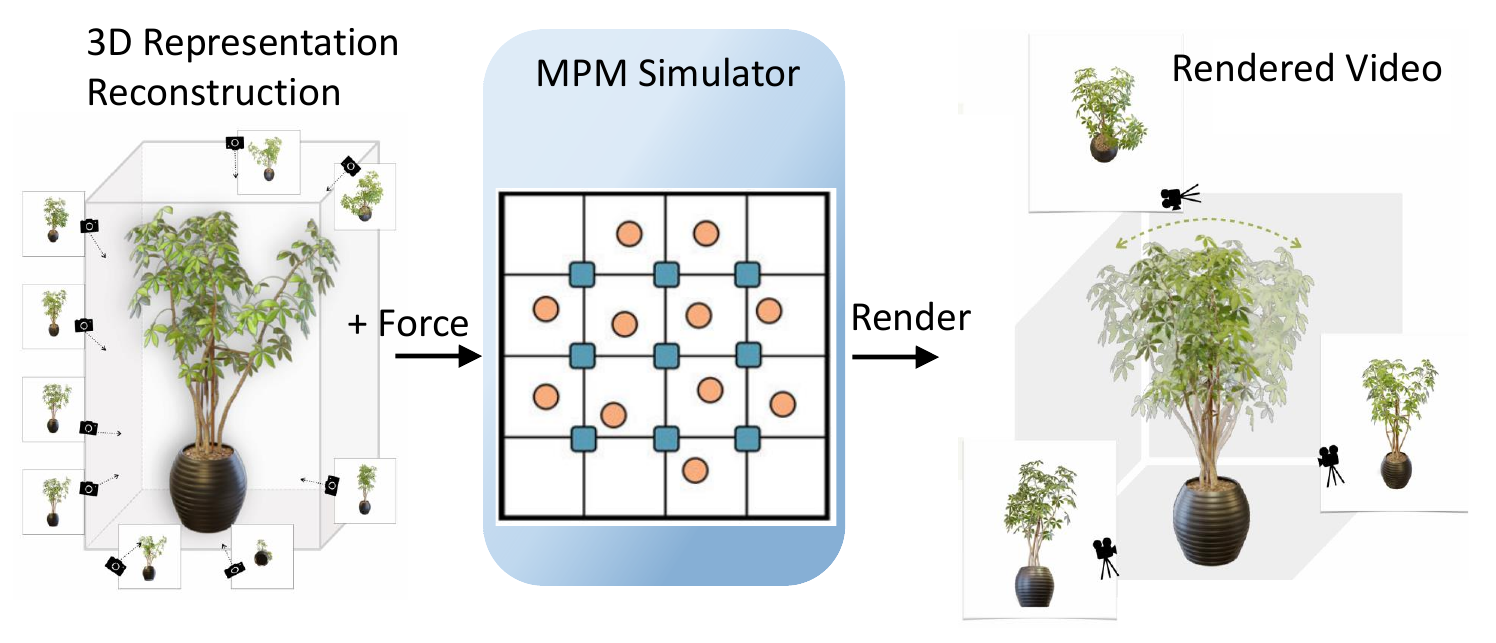}
    \caption{Generation pipeline based on physical simulation, cf. Physguassian\cite{xie2024physgaussian}.}
    \label{fig:phys_sim_pipeline}

\end{figure}

\subsubsection{Interactive dynamic generation}\label{sec:interactive_dynamic}

Interactive dynamic generation utilizes external forces as inputs to static 3D scenes, employing physics simulators to simulate the interactions between the scene and the applied forces, infer the evolution trends that comply with physical laws, and generate corresponding dynamic scenes.

\textbf{Dynamic Generation Based on GS.} To capture realistic motion patterns and causal relationships for creating more lifelike interactive dynamics, PhysGaussian\cite{xie2024physgaussian} was the first to integrate 3D Gaussian Splatting (3DGS) with the Material Point Method (MPM) into a unified simulation-rendering pipeline, as shown in \cref{fig:phys_sim_pipeline}. This pioneering approach treats 3D Gaussian kernels as discrete particles, allowing the deformation of Gaussian kernels during continuous media transformation to seamlessly integrate physical simulation with visual rendering. Compared to PhysGaussian\cite{xie2024physgaussian}, VR-GS\cite{jiang2024vr} addresses the real-time inefficiencies of MPM by adopting XPBD\cite{macklin2016xpbd}. Additionally, it introduces a two-stage embedding strategy to resolve sharp artifact issues: Gaussian kernels are first embedded into local tetrahedra independently, and the tetrahedral vertices are then embedded into a global grid, allowing the gaussian kernels to adapt smoothly to the mesh. Gaussian Splashing (GSP)\cite{feng2024gaussian} innovatively integrates Lagrangian fluid-solid interactions within 3DGS scenarios through a unified PBD framework. Notably, GSP decouples the simulation and rendering processes at the solid level through separate instantiation of particles and Gaussian kernels. Furthermore, it enhances fluid simulation and rendering by optimizing Gaussian kernels with the integration of diffuse reflection, normals, specular highlights, and surface roughness.  Physmotion\cite{tan2024physmotion} and Phy124\cite{lin2024phy124} focus on generating physics-consistent dynamic videos from a single image. Sync4D\cite{fu2024sync4d} maps the motion from the reference video onto a skeleton with skinning weights, and integrates MPM-based physical simulation and displacement loss to optimize the velocity field.

\textbf{Dynamic Generation Based on NeRF.} In addition to 3DGS-based physical simulation, PIE-NeRF\cite{feng2024pie} and  Video2Game\cite{xia2024video2game} integrate NeRF with physics simulation. PIE-NeRF demonstrates the feasibility of integrating classical Lagrangian dynamics with a mesh-free NeRF approach, introducing Quadratic Generalized Moving Least Square (Q-GMLS) and Voronoi partitioning to handle nonlinear deformations and intensive computations. Video2Game\cite{xia2024video2game} employs NeRF to construct 3D worlds for games, further refining mesh models to enhance compatibility with game engines.

\textbf{Neural Networks for Solving PDEs.} Beyond incorporating physics simulators to constrain and guide physical video generation, some works (e.g., ElastoGen\cite{feng2024elastogen}) embed explicit physical priors into neural models. During training, these methods optimize for the numerical solutions of partial differential equations (PDEs), enabling collaborative learning between physics and neural networks.

\begin{figure}[t]
    \centering
    \includegraphics[width=1\linewidth]{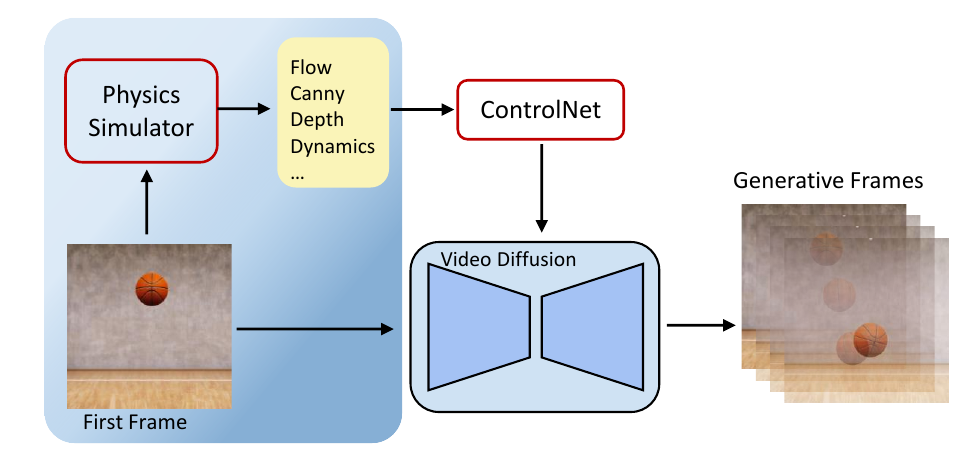}
    \caption{Physics simulator generates motion signals.}
    \label{fig:phys_signal}
\end{figure}
\subsubsection{Physicalization of Motion Field} \label{sec:phys_motion}
While physics simulation-based methods(see \cref{sec:interactive_dynamic}) can faithfully reconstruct dynamic scenes that adhere to physical laws, they are limited in scope and cannot adapt to the diversity of real-world scenarios. In contrast, the motion-guided generation paradigm (see \cref{sec:motion_guide}) offers a certain degree of flexibility but relies on manually defined motion signals, often leading to inaccuracies due to weak geometric understanding and missing physical constraints. To this end, the academic community proposes a new framework (see \cref{fig:phys_signal}) -- before the generation stage of the video diffusion model, first construct motion signals that conform to physical laws (such as optical flow fields, depth maps, dynamic assets, etc.), and encode these physical constraint conditions into the generation process of the diffusion model, thereby achieving precise control of video synthesis.

To optimize the generation of motion signals, MotionCraft\cite{aira2024motioncraft} introduces a physics-informed zero-shot video generation approach that achieves dual-consistency motion mapping between latent and pixel spaces by strategically warping noise latent vectors through physically simulated optical flow. Similarly, PhysAnimator\cite{xie2025physanimator} innovatively integrates physical simulation with data-driven generative models to enable the generation of deformable animation sequences. This approach solves the kinematic equations in 2D space to evolve dynamics and generate conditional optical flow fields. The optical flow fields are then used to generate deformable sketch sequences as control signals, with sampled keyframes guiding the coherent animation generation. GPT4Motion\cite{lv2024gpt4motion} utilizes GPT-4 to generate Blender scripts that simulate continuous motion sequences, producing depth and edge maps. These outputs serve as conditional inputs for ControlNet, guiding a diffusion model to render realistic videos frame by frame. Similarly, PhysGen\cite{liu2025physgen} employs rigid-body physics and large language model-inferred parameters to simulate realistic dynamic interactions, driving a video diffusion module for rendering and refinement. Physmotion\cite{tan2024physmotion} uses physically plausible foreground dynamics generated by MPM-simulated 3DGS as intermediate coarse 3DGS dynamics, and then constructs a diffusion pipeline employing DDIM+ inversion to obtain latent noise codes. During the sampling stage, it mixes coarse and enhanced sampled videos to achieve high-quality results. 

Moreover, using such above simulations as physical priors to guide diffusion models in video generation may introduce discrepancies between simulated and real-world conditions (Sim2Real Gap). To address this issue, SimGen\cite{zhou2024simgen} proposes a cascaded generation framework, where a simulator first generates simulated conditions (e.g., depth and segmentation) combined with text prompts, which are then fed into a lightweight diffusion model to transform them into realistic conditions. These refined conditions subsequently guide the diffusion model in generating realistic driving scenes.

\begin{table*}[h]
\centering
\renewcommand{\arraystretch}{1.5}
\resizebox{\textwidth}{!}{%
\begin{tabular}{m{1.8cm}<{\centering}m{2.5cm}<{\centering}m{2cm}<{\centering}m{3cm}<{\centering}m{1.9cm}<{\centering}m{0.8cm}<{\centering}m{1.5cm}<{\centering}m{1.5cm}<{\centering}m{3cm}<{\centering}}
\toprule
&\multirow{3}{*}{Methods } & \multirow{3}{*}{Input Type} & \multirow{3}{*}{Physics Simulator } & \multicolumn{3}{c}{Material Field} &   \multirow{3}{*}{Representation} & \multirow{3}{*}{Materials Types}\\ \cmidrule(lr){5-7}
& &&& Manual Parameter init. & Learnable & LLM inference& \\ \midrule
 
\multirow{9}{*}{\makecell[c]{Interactive \\Dynamic \\ Generation}}& PhysGaussian\cite{xie2024physgaussian}& Multi-view & MPM & $\checkmark$ &  &  & 3DGS & varieties materials \\ 
&Phy124\cite{lin2024phy124}& Single image& MPM & $\checkmark$ &  &  & 3DGS & elastoplasticity\\ 
&VR-GS\cite{jiang2024vr}& Multi-view& XPBD & $\checkmark$ &  &  & 3DGS & elastoplasticity \\ 
&Gaussian Splashing\cite{feng2024gaussian}& Multi-view& PBD & $\checkmark$ &  &  & 3DGS &solids and fluids \\ \cmidrule(lr){2-9}

&PIE-NeRF\cite{feng2024pie}& Multi-view & Q-GMLS/Taichi & $\checkmark$ &  &  & NeRF& hyperelastic\\ 
&Video2Game\cite{xia2024video2game}&Dynamic Video & Cannon.js/Blender/Unreal
 & $\checkmark$ &  &$\checkmark$  & NeRF& Rigid-body \\ 
&ElastoGen\cite{feng2024elastogen}& 3D model & NeuralMTL & $\checkmark$ &  &  & NeRF/ 3DGS& hyperelastic \\ \midrule

\multirow{5}{*}{\makecell[c]{Physicalization \\of Motion \\Field}}&PhysGen\cite{liu2025physgen}& Single image & Pymunk &  &   & $\checkmark$ & 2D & rigid-body \\
&MotionCraft\cite{aira2024motioncraft}& Text & $\phi$-Flow  & $\checkmark$ &   & & 2D & rigid-body and Fluids\\
&PhysMotion\cite{tan2024physmotion} & Single image & MPM & $\checkmark$ &   & & 3DGS \& 2D & varieties materials\\
&GPT4Motion\cite{lv2024gpt4motion} & Text & Blender &  &   & $\checkmark$& 2D & varieties materials \\
& PhysAnimator\cite{xie2025physanimator}&Single anime illustration & Taichi\cite{hu2020difftaichi} & $\checkmark$&&& 2D&deformable body\\
\midrule
\multirow{13}{*}{\makecell[c]{Material\\ Space\\ Simulation }}&PhysDreamer\cite{zhang2025physdreamer}&3D model & MPM &  & $\checkmark$ &  & 3DGS& hyperelastic  \\ 
&PAC-NeRF\cite{li2023pac}& Dynamic Video & MPM &  & $\checkmark$ &  & NeRF& varieties materials\\ 
&Physics3D\cite{liu2024physics3dlearningphysicalproperties}&3D model & MPM &  & $\checkmark$ &  & 3DGS& elastoplastic and viscoelastic  \\ 
&DreamPhysics\cite{huang2024dreamphysics}&3D model \& Text \& Image & MPM &  & $\checkmark$ &  & 3DGS& elastoplastic  \\ 
&Liu et al.\cite{liu2024unleashing}&Multi-view \& Text & MPM &  & $\checkmark$ & $\checkmark$ & 3DGS& varieties materials \\ 

&NeuMA\cite{caoneuma}&Multi-view & MPM &  & $\checkmark$ & & 3DGS& varieties materials  \\ 
&OmniPhysGS\cite{anonymous2024omniphysgs}&3D model \& Text & MPM &  & $\checkmark$ &  & 3DGS& varieties materials  \\ 
\cmidrule(lr){2-9}
&Feature Splatting\cite{qiu2024featuresplattinglanguagedrivenphysicsbased}&Multi-view \& Text & GS-Taichi-MPM &  &  &$\checkmark$  & 3DGS& varieties materials  \\ 
&Phys4DGen\cite{lin2024phys4dgen} & Singe image & MPM &  &  &$\checkmark$  & 3DGS& varieties materials\\ 
&Sim Anything\cite{zhao2024automated} & Multi-view & MLS-MPM &  &  &$\checkmark$  & 3DGS& varieties materials \\ 
& GaussianProperty\cite{xu2024gaussianproperty}&Multi-view&MPM&&&$\checkmark$ &3DGS&varieties materials\\
 \bottomrule
\end{tabular}
}
\vspace{5pt}
\caption{Summary of physics simulation-based generation methods.}
\vspace{-14pt}
\end{table*}
\subsection{Material Space Simulation}\label{sec:material}
Interactive dynamic generation (see \cref{sec:interactive_dynamic}), based on rigorously defined mathematical equations and physical simulators, generates high-precision interactive dynamics\cite{xie2024physgaussian,feng2024pie,xia2024video2game} with physical laws through numerical simulations. However, the interpretability of this approach heavily relies on manually defined physical parameters (e.g., Young’s modulus, friction coefficients)\cite{liu2024physics3dlearningphysicalproperties,zhang2025physdreamer}. These parameterized representations are constrained by parameter sensitivity and diversity in real-world scenarios, making it challenging to configure complex material fields in scenes. To address these limitations, current research utilizes various physical priors to guide the adaptive modeling of material fields. By autonomously learning and inferring material fields, these methods\cite{liu2024physics3dlearningphysicalproperties,huang2024dreamphysics,li2023pac,zhang2025physdreamer,caoneuma,lin2024phys4dgen} enable more accurate and adaptable physical simulations.

\begin{figure}
    \centering
    \includegraphics[width=1\linewidth]{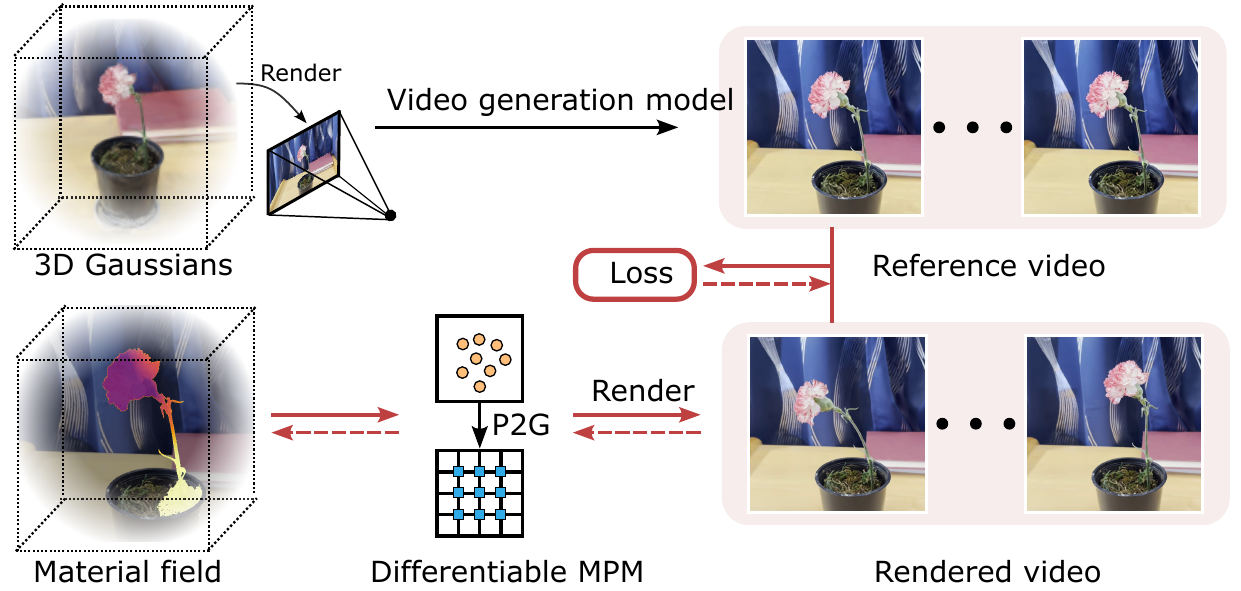}
    \caption{Architecture of PhysDreamer\cite{zhang2025physdreamer}. Optimize the material field and velocity field by minimizing the discrepancy between the rendered video and the reference video.}
    \label{fig:fig_physdreamer}
\end{figure}

\textbf{Score Distillation Sampling.} To enable the automatic estimation of material properties, DreamFusion\cite{poole2022dreamfusion} introduced Score Distillation Sampling (SDS), which provides
a novel approach to bridging the gap between 2D and 3D representations. SDS leverages the score function of diffusion models to optimize the parameters of 3D models (e.g., NeRF or other neural implicit representations) via
backpropagation. By minimizing a noise reconstruction loss in an adversarial manner, SDS ensures that the 2D-rendered images generated by the 3D representation align with the distribution of the diffusion model, typically conditioned on textual guidance for controlled generation.

\begin{equation}
    \nabla_{\theta} \mathcal{L}_{\text{SDS}} = \mathbb{E}_{\epsilon, t} \left[ w(t) \cdot \left( \epsilon_{\theta} (\mathbf{x}_t) - \epsilon \right) \cdot \nabla_{\theta} \mathbf{x}_0 \right]
\end{equation}
where $\epsilon_{\theta}$ represents the noise predictor of the diffusion model, $\mathbf{x}_t$ is the noisy sample, and $\mathbf{x}_0$ is the current rendered
image.

Through SDS, researchers can leverage existing 2D diffusion models to optimize differentiable material fields directly. Physics3D\cite{liu2024physics3dlearningphysicalproperties} extends the physical parameters in MPM to capture both the elastic and viscous properties of materials. DreamPhysics\cite{huang2024dreamphysics}, based on the physically modeling-friendly KAN (Kolmogorov Arnold Networks)\cite{liu2024kan} representation, introduces Motion Distillation Sampling (MDS) to emphasize motion information in videos, ensuring temporal consistency. MDS builds on SDS by reducing biases caused by color and placing greater focus on motion. However, applying multiple iterations of SDS using a diffusion model results in a significant increase in computational overhead. Constrained by the computationally intensive requirements of SDS, Liu et al.\cite{liu2024unleashing} proposes a resource-efficient optical flow loss as an alternative to optimize material properties. 

\textbf{Learning from the Reference Video.} PAC-NeRF\cite{li2023pac} introduces a hybrid particle and grid-based NeRF representation and estimates the unknown geometry and physical parameters of dynamic objects in a supervised manner. PhysDreamer\cite{zhang2025physdreamer} (see \cref{fig:fig_physdreamer}) distills physics priors via aligning a reference video from video generation models, modeling and optimizing physical material fields (e.g., Young’s modulus and Poisson's ratio) while incorporating MPM for material dynamics simulation. Different from the previous method of treating 3D particles equally, FluidNexus\cite{gao2025fluidnexus} divides 3D particles into physical particles and visual particles, and uses a differentiable physical simulator to reconstruct and predict physical particles and synthesize new perspective fluid videos to optimize the visual appearance of 3D fluids.

\begin{figure*}[ht]
    \centering
    \includegraphics[width=0.9\linewidth]{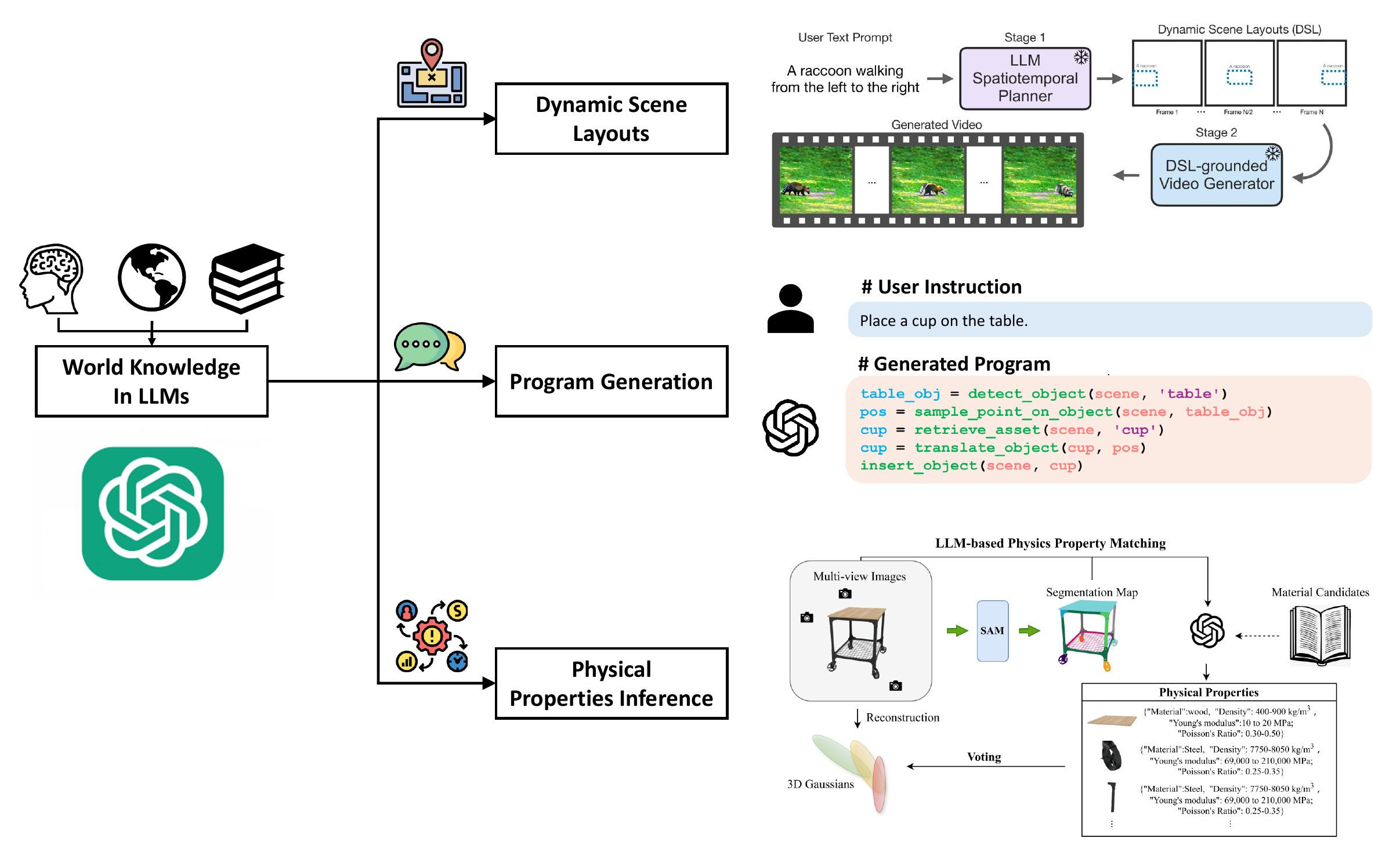}
    \vspace{-10pt}
    \caption{World physics knowledge in LLMs is used for scene layouts, program generation, and properties inference.}
    \label{fig:llm_apply}
\end{figure*}

\textbf{Learning Constitutive Models.} Compared to approaches that model individual materials, methods capable of handling a wide range of materials and complex objects are evidently more aligned with the real world. OmniPhysGS\cite{anonymous2024omniphysgs} extends gaussian kernels into learnable constitutive Gaussian kernels and predicts material properties using expert-designed constitutive models, enabling flexible adaptation to diverse materials without the need for manual configuration. On the other hand, Physics3D\cite{liu2024physics3dlearningphysicalproperties} adopts inelastic and dissipative terms in its constitutive model, which is limited to modeling a few specific physical quantities. NCLaw\cite{ma2023learning} innovatively supplants classical elastoplastic constitutive modeling with a neural network-based approach, strategically embedding data-driven constitutive laws within a differentiable, partial differential equation (PDE)-governed simulation framework. The key idea behind NeuMA\cite{caoneuma} is to learn the discrepancy between expert physical models and real-world scene dynamics, and perform personalized residual correction. Specifically, it uses the NCLaw\cite{ma2023learning} encoded with universal material priors, combined with the efficient, adaptive learning of elasto-plastic properties via the Lora adapter. Additionally, unlike PhysGaussian, which directly discretizes Gaussian kernels into particles, NeuMA samples physical particles to drive the differentiable rendering of Gaussian kernels.

\textbf{LLMs Inference Physics Properties.} Additionally, the natural physical knowledge and reasoning abilities embedded in LLMs open up new possibilities for material reasoning in 4D dynamic generation. These models can capture complex physical phenomena and material properties from images or even videos. We will provide a detailed introduction to the methods involving LLMs for reasoning about physics properties in \cref{sec:llm_sim}, such as Feature Splatting\cite{qiu2024featuresplattinglanguagedrivenphysicsbased}, Phys4DGen\cite{lin2024phys4dgen}, Sim Anything\cite{zhao2024automated}, GaussianProperty\cite{xu2024gaussianproperty}.

\subsection{LLMs Empowering Physical Simulation}\label{sec:llm_sim}
Although the methods discussed above provide explicit strategies for spatiotemporal\cite{wang2024levitor,bahmani2025tc4d,niu2025mofa} layout modeling or material field distillation\cite{liu2024physics3dlearningphysicalproperties,anonymous2024omniphysgs,zhang2025physdreamer}, they often rely on complex priors, which can reduce generalizability. In contrast, LLMs demonstrate remarkable capability in memorizing world knowledge due to their vast model capacity and extensive training datasets\cite{min2023recent}. This enables LLMs to recall relevant physical knowledge to provide contextually appropriate responses across diverse scenarios\cite{yu2025stochastic,mitchell2023debate}. Consequently, recent studies have sought to leverage LLMs to enable more efficient spatiotemporal reasoning and automated material property estimation, thereby generating high-quality videos, as shown in \cref{fig:llm_apply}.

\textbf{Dynamic Scene Layouts.} LLM-Grounded Video Diffusion (LVD)\cite{lian2024llmgrounded} capitalizes on the remarkable natural language understanding and reasoning capabilities of LLMs to generate physically plausible spatiotemporal scene layouts, eliminating the need for additional motion priors by using cross-attention alignment. C3V\cite{zhucompositional} provides coarse 3D object trajectories estimated by LLMs and employs video diffusion models for fine-grained supervision. Trans4D\cite{zeng2024trans4d} uses MLLMs to generate scene descriptions that include physical properties and dynamic spatiotemporal information for initializing 4D scene modeling. It then employs a geometric transition network to predict intermediate scenes of complex object interactions from coarse to fine levels. 

\textbf{Program Generation.} Although LLMs are generally recognized for their capability to understand physical common sense, they face significant limitations in reasoning about dynamic real-world interactions. To overcome this, Kubrick\cite{he2024kubrick} focuses on generating videos with physical correctness, precise camera control, and temporal consistency in an end-to-end manner. Built on an RAG framework, it uses LLM/VLM agents along with 3D engines (e.g., Blender) to design an LLM Director, LLM Programmer, and LLM Reviewer, enabling iterative refinement of synthetic videos. Similarly, both AutoVFX\cite{hsu2024autovfx} and GPT4Motion\cite{lv2024gpt4motion} integrate the physics engine Blender with the large language model GPT-4\cite{achiam2023gpt}. AutoVFX enables users to create executable programs through natural language instructions, driving dynamic editing and rendering of 3D modeling scenes. GPT4Motion, on the other hand, embeds components simulated by the physics engine (such as edge maps and depth maps) as conditional signals into the Stable Diffusion model to generate video frames with physically plausible motion. LLMPhys\cite{cherian2024llmphy} combines LLMs with physics engines for collaborative reasoning of dynamic scene changes. Specifically, LLMs infer physical parameters and generate simulation programs, iteratively refining reasoning accuracy through feedback. The inferred parameters are then used in simulation tasks to produce complete dynamic sequences. 

\textbf{Physical Properties Generation.} LLMs and other large pretrained models also play a pivotal role in material-level reasoning for dynamic generation. When dealing with complex interactions in 4D scenes, Feature Splatting\cite{qiu2024featuresplattinglanguagedrivenphysicsbased} manipulates object appearance and assigns material properties through natural language guidance. Then, the method extends MPM using GS-Taichi-MPM by integrating Gaussian-specific features (e.g., isotropic opacity and covariance) to address issues such as collision collapse and artifacts. Similar to PhysDreamer\cite{zhang2025physdreamer}, Phys4DGen\cite{lin2024phys4dgen} focuses on material-centric modeling in constructing 4D dynamic spaces. While PhysDreamer\cite{zhang2025physdreamer} models physical material fields via neural representations, Phys4DGen leverages large pre-trained models to segment and infer material properties. %
PhysGen\cite{liu2025physgen} leverages GPT-4o\cite{achiam2023gpt} to automatically assign physical parameters to physically accurate motion fields, enhancing video generation with diffusion models. GaussianProperty\cite{xu2024gaussianproperty} utilizes GPT-4V to estimate material properties, applying a voting strategy to project physical attributes onto 3D Gaussian representations for realistic dynamic simulations. It also leverages these material properties for Robot grasp prediction, ensuring objects remain intact by avoiding excessive deformation during manipulation. Similarly, Sim Anything\cite{zhao2024automated} relies on MLLMs to predict the overall material properties of objects and reformulates local material property variations as probabilistic distribution estimates, aiming to capture a comprehensive material distribution for realistic physical simulation. 

\textbf{Discussion.} Overall, passive physical cognition-based generation relies on pre-stored physical knowledge, such as physics simulators, symbolic representations, or LLMs, to enhance the physical plausibility of generated videos. While this approach improves physical interpretability and consistency, it faces several fundamental limitations: 1) Limited Adaptability to Unseen Scenarios: These models passively retrieve and apply predefined physical knowledge, restricting their ability to generalize beyond observed physical phenomena or adapt to novel interactions and environmental conditions. 2) When relying on LLMs or other symbolic representations, there is often a gap between theoretical physical principles and their practical applicability in real-world video generation tasks. 3) Integrating high-fidelity physics simulators often incurs significant computational costs, making real-time video generation a challenging task. To address these limitations, future research could explore the following directions: (i) incorporating active interaction with the environment to allow models to dynamically refine physical world understanding, enhancing both generalization and adaptability; and (ii) developing more efficient and differentiable physics simulators that can be seamlessly integrated with world models to improve both computational efficiency and physical fidelity in video generation.

\section{Active Cognition for World Simulation}\label{sec:active_cognition}
In the development of cognitive architectures, physical symbol-driven generative systems (e.g., physics engine-based simulations) operate under strict rules but often fail to capture the diverse physical phenomena present in complex real-world scenarios, limiting their predictive power in open-ended environments(see \cref{sec:passive_cognition}). In contrast, a key advantage of world models lies in their ability to infer potential scenarios beyond the training data distribution, thereby enabling diverse world simulations for embodied agents.
As envisioned by LeCun, the ultimate goal of world models is to reason and plan for unknown patterns, understanding and predicting the evolution of the world in a human-like manner \cite{lecun2022path}. All these analyses highlight the necessity for world models to incorporate an imagination mechanism grounded in physical commonsense. Consequently, a critical challenge for developers is how to integrate real-world physical cognition into the design and learning of world models, enabling a profound understanding of the physical world's dynamics and ensuring that future predictions are physically faithful. To achieve this, world models may dynamically update through active environmental interaction, bridging the gap between simulation and real-world complexity. 

\begin{figure*}
    \centering
    \includegraphics[width=1\linewidth]{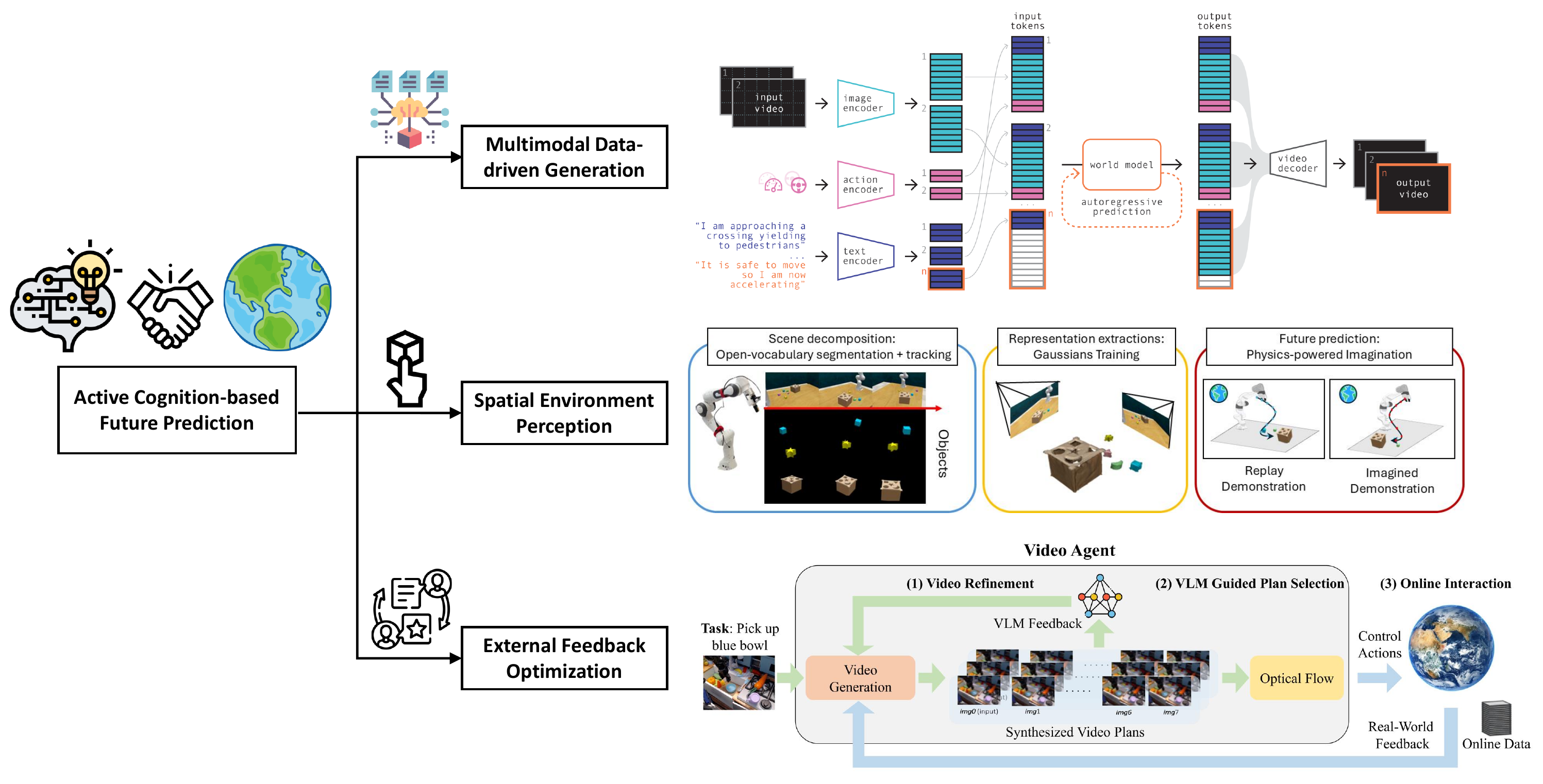}
    \caption{The model actively interacts with the environment to achieve future prediction through multimodal data-driven generation, spatial environment perception, and external feedback optimization.}
    \label{fig:interact_world}
\end{figure*}

\subsection{Multimodal Data-driven Generation} \label{sec:multimodal}

Currently, most world models learn statistical representations from large-scale datasets and are gradually showing their potential as foundational physical simulators. World models need to integrate multimodal inputs, such as visual, linguistic, and action information, in order to construct a comprehensive representation of the environment and enable planning in the latent space (similar to how humans perceive the world through a combination of sight, sound, and touch).

Building on OpenAI's remarkable success in LLMs\cite{radford2019language,brown2020language,achiam2023gpt}, Sora\cite{sora} aims to achieve AGI (Artificial General Intelligence) in the vision domain and emergent intelligence at the physical level by significantly increasing the scale of training data and parameters. Sora stands out with its ability to generate high-quality, controllable, and realistic videos up to one minute long. Its introduction marks a critical step in the evolution from generative models to world models, inspiring further advancements in this field. However, Sora faces significant challenges in accurately simulating physical processes and interactions, and merely scaling data has yet to fulfill the envisioned higher-level intelligence of world models\cite{zhu2024sora}.

\textbf{Interactive.} Genie\cite{bruce2024genie}, a contemporary of Sora, focuses on creating interactive environments and controllable virtual worlds, offering users diverse interaction experiences. By combining implicit action models, video tokenizers, and dynamic models, Genie extracts latent actions, compresses video frames into discrete tokens, and predicts subsequent video tokens through the dynamics model. Genie 2\cite{genie2} extends this capability into 3D scenarios, generating dynamic virtual scenes from textual prompts or images. Leveraging Imagen 3\cite{baldridge2024imagen}, an advanced text-to-image model, and incorporating Classifier-Free Guidance (CFG), Genie 2 achieves flexible and precise scene generation, simulating physical effects such as fluid dynamics, smoke, lighting, gravity, and more. UniSim\cite{yang2024learning} recognizes ``interacting with the environment'' as a key trend and introduces a universal world-generation simulator conditioned on actions. It unifies various action modalities (e.g., language instructions, robot controls, and camera movements) into a shared action space, which serves as conditional prompts to guide a diffusion model in generating subsequent frames. UniSim provides a unified environment for training both high-level vision-language policies and low-level reinforcement learning strategies, bridging the gap between simulated and real-world environments.
To enhance generative models' causal control and intervention capabilities in the physical world, WorldDreamer\cite{wang2024worlddreamer} proposes a novel approach using vision-text-action triplet datasets for supervised training. Its U-ViT-based\cite{hoogeboom2023simple} STPT (Spatial Temporal Patchwise Transformer) integrates dynamic masked strategies, embedding visual and prompt information to generate results in approximately 10 steps, 3 to 20 times faster than diffusion models.

The application of generative world models in autonomous driving has become a widely recognized research direction. In particular, autonomous driving\cite{hu2023gaia1generativeworldmodel,gao2024vista,wang2023drivedreamer} requires safe and reliable decision-making in unstructured and complex scenarios. Pioneering work, such as GAIA-1\cite{hu2023gaia1generativeworldmodel}, encodes past images, text, and actions to predict the next image token, emphasizing a deep understanding of driving data. Building on this, DriveDreamer\cite{wang2023drivedreamer} introduces a two-stage generation strategy: in the first stage, it integrates various driving elements to comprehensively understand underlying traffic structures and construct accurate scene maps; in the second stage, it predicts future video frames to enable controllable generation of interactive driving scenarios. Unlike GAIA-1, which focuses on scene video generation, DriveDreamer places greater emphasis on the decision-making aspects of generating driving behaviors and scenes.

\textbf{Digitalized Physical World.} Despite these advances, Kang et al.\cite{kang2024farvideogenerationworld} remain skeptical about the ability of vision-based world models to abstract physical laws. Studies involving diverse physical scene datasets evaluated models' capabilities under in-distribution, out-of-distribution, and compositional generalization settings. Even with aggressive scaling, models struggled to reproduce correct physical phenomena, merely memorizing and imitating observed physics dynamics. Similarly, Motamed et al.\cite{motamed2025generative} proposes to evaluate various general video generation models on the Physics-IQ dataset, and the conclusions obtained are similar to the view of Kang et al.\cite{kang2024farvideogenerationworld}. These views indicate that vision-based representations alone are insufficient for precise physical modeling.
Cosmos\cite{agarwal2025cosmos} seeks to digitize the physical world with the goal of creating ``Physical AI'', leveraging generalist generative models and physical simulators to design controlled scenarios. These scenarios are employed to fine-tune the model's understanding of physical attributes such as gravity, collisions, torque, and inertia. Cosmos enables generate high-quality, 3D-consistent videos while advancing physical AI tasks. The World Simulation Assistant (WISA)\cite{wang2025wisa} constructs a dataset of 32,000 videos with captions to fine-tune T2V models. It decomposes physics-based captions into textual descriptions, qualitative categories, and quantitative attributes for multidimensional interpretation. By integrating a Mixture-of-Physical-Experts Attention (MoPA) mechanism and a Physics Classifier, WISA enables independent understanding of physical properties. This structured approach effectively embeds physical principles into the model, enhancing its ability to generate physics-consistent videos. PISA\cite{li2025pisaexperimentsexploringphysics} focuses on physics-based generation in free-fall scenarios, exploring a two-stage post-training strategy using Open-Sora\cite{zheng2024open} as the foundation model, incorporating physics-supervised fine-tuning and object reward optimization. Evaluation results demonstrate that fine-tuning an open-source model on a small dataset with this approach enables it to acquire new capabilities for generating more physically accurate videos.

\begin{table*}[ht]
\renewcommand{\arraystretch}{1.5}
    \centering
    \resizebox{\textwidth}{!}{
    \begin{tabular}{m{2cm}<{\centering}m{3cm}<{\centering}m{3cm}<{\centering}m{3cm}<{\centering}m{5cm}<{\centering}}
    \toprule
        Methods&Reward Model&Feedback Type& Optimization Technique&Optimization Direction\\ \midrule
        IPO\cite{yang2025ipo} &Human-Annotated Training VLM  & Pair-wise \& Point-wise feedback& Diffusion-DPO\cite{wallace2024diffusion} \& Diffusion-KTO\cite{li2024aligning} & Subject consistency, motion smoothness and aesthetic quality\\ 
         VideoReward\cite{liu2025improving}&Human-Annotated Training VLM  & Pair-wise feedback & Flow-DPO\& Flow-RWR\&Flow-NRG & Visual quality, motion quality, and text alignment\\
         Furuta et al.\cite{furuta2024improving}& Gemini-1.5-Pro\cite{team2023gemini} \& Metric & Pair-wise \& Point-wise feedback & RWR]\cite{peters2007reinforcement} \& DPO\cite{rafailov2023direct} & Overall coherence, physical accuracy, task completion, and the existence of inconsistencies\\
         VideoAgent\cite{soni2024videoagent}& GPT4-turbo\cite{achiam2023gpt} \& Online Execution Feedback & Binary value $\{0,1\}$ & Consistency models\cite{song2023consistency} \& Online finetuning&Trajectory smoothness, physical stability and achieving the goal \\
         Gen-Drive\cite{huang2024gen}& GPT-4o\cite{achiam2023gpt} & Pair-wise feedback  & DDPO\cite{black2023training}  & Complex traffic environment, scene consistency and interactive dynamics\\
         PhyT2V\cite{xue2024phyt2v}& GPT-4o\cite{achiam2023gpt} & Mismatch between video semantics and prompts & LLM global step-back reasoning & Adherence physical rules \\
        \bottomrule
    \end{tabular}
    }
    \vspace{10pt}
    \caption{Overview of the characteristics of various methods in external feedback optimization.}
    \vspace{-13pt}
    \label{tab:feedback_methods}
\end{table*}

\subsection{Spatial Environment Perception}\label{sec:spatial_env}
Traditional world models typically rely on semantic representations and lack scene-level spatiotemporal dynamic modeling, which may result in poor performance in interactive tasks. Especially in embodied environments, robots need to execute precise spatiotemporal actions, and relying solely on semantic abstractions may lead to physically infeasible decisions. Spatiotemporal perception-driven world models address the limitations of multimodal data-driven approaches in the spatial perception dimension, physical consistency, and the bottleneck of diverse data imagination, by leveraging structured spatiotemporal representations.

\textbf{Spatial Perception.} Spatial perception, based on geometric priors, enables 3D scene reconstruction, overcoming the inherent planar bias of traditional 2D video generation.
ManiGaussian\cite{lu2025manigaussian} extends this approach by predicting robotic actions from geometric, semantic, and dynamic perspectives, propagating Gaussian particles over time to capture spatiotemporal scene dynamics. It supports language-controlled agents by dynamically propagating semantic features within a Gaussian distribution. Additionally, it constructs a Gaussian world model to parameterize the dynamic Gaussian splatting framework, leveraging real-world scenes to supervise the learning of Gaussian deformation fields, thereby predicting future scenarios. 
Realistic driving video generation must also adhere to fundamental physical laws, such as absolute and relative motion and spatial relationships. The key insight of DrivePhysica\cite{yang2024drivephysica} is aligning ego-vehicle coordinates with global world coordinates to achieve complementary perspectives while capturing the relative motion of surrounding objects to generate instance flows. By embedding 3D bounding box coordinates, DrivePhysica incorporates depth information to preserve correct spatial relationships. By integrating critical physical principles, DrivePhysica generates high-quality, multi-view driving videos.

\textbf{Physics Consistency.}
Physics consistency constraint modeling explicitly encodes the physical laws in space, providing a corrigible framework of physical constraints for feedback regulation in the perception, planning, and control modules, ensuring that the model's decisions align with the action logic of the real world. Abou-Chakra et al.\cite{abou-chakra2024physically} models robotic environments as 3D Gaussian distributions representing visual states and introduces position-based dynamics (PBD)\cite{muller2007position} simulation. This enables the model to predict future states and perform real-time corrections under strict physical constraints. By embedding explicit physical priors (particles) into the 3D Gaussian representation, the model utilizes visual feedback to adjust Gaussian distributions, thereby refining particle positions. This approach allows robots to robustly understand physical laws and synchronize physical simulation with visual feedback, advancing perception, planning, and control algorithms.

\textbf{Diverse Data Imagination.}
Most of the aforementioned methods require a large volume of real-world samples; however, in practice, embodied data is often scarce. Developing a world model capable of comprehensive learning under limited sample conditions has become a significant challenge. DreMa\cite{barcellona2024dream} introduces an innovative and valuable compositional manipulative world model designed to create diverse realistic environments (accommodating the motion dynamics and physical properties) for robots. By leveraging generative simulations and physical simulators, DreMa generates physically plausible, novel, and imagined dynamic demonstrations (with specific equivariant transformations applied). This enables robots to achieve significant improvements in imitation learning with limited data. By extending the applicability of world models beyond training domains, DreMa narrows the gap with real-world environments. Moreover, reconstructing the complexity of dynamic interactive driving scenes remains a significant challenge. Considering the natural advantages of generative world models in producing diverse and controllable high-fidelity 2D videos, DriveDreamer4D\cite{zhao2024drivedreamer4d} innovatively leverages world model priors to advance autonomous driving 4D scene reconstruction. Specifically, it adjusts original trajectories to generate new trajectories, guiding the world model to produce diverse dynamic data (e.g., lane changes, acceleration, and deceleration). Then, it integrates real and synthetic data to optimize the performance of the 4D scene generation model.

\subsection{External Feedback Optimization}\label{sec:feedback}
External Feedback-based video generation dynamically adjusts the model’s generation of the world by incorporating external environmental knowledge or real-time environmental signals. The core idea is to leverage feedback information from sources external to the model’s own training data to ensure that the generated content aligns with specific domain physical laws, semantic logic, or human preferences, thereby achieving an active cognitive closed-loop optimization of “generation-environment observation-correction update.” 
\Cref{tab:feedback_methods} presents the implementation methods and characteristics of different external supervision feedback approaches.

\textbf{Training the Reward Model through Human Preference Annotations.} To address the physical illusions in dynamic interactive scenes of video generation, the Iterative Preference Optimization (IPO)\cite{yang2025ipo} framework trains a reward model that automatically evaluates the physical plausibility of generated videos and then iteratively optimizes the generation model based on these physics-based preference feedbacks.
VideoReward\cite{liu2025improving} proposed a large-scale human preference dataset consisting of 182k annotations, covering human evaluations of video generation in terms of visual quality, motion quality, and text alignment. This dataset is used to train the VLM reward model, where multidimensional reward tokens are decoupled to maintain contextual independence across evaluation levels. Finally, through exploring alignment algorithms-two training-time strategies (Flow-DPO\cite{wallace2024diffusion} and Flow-RWR\cite{peng2019advantage}) and one inference-time technique (reward-guided)-video alignment is achieved.

\textbf{Replacing Human Feedback with AI Models.} However, the cost of manual annotation is high. To solve this problem, some methods replace manual annotation with AI agents to obtain human preference feedback. Furuta et al.\cite{furuta2024improving} combines RL fine-tuning strategies to accept external feedback and iteratively optimize object motion to match the real world. This approach explores feedback mechanisms based on metrics from complex motion scenarios, as well as AI-driven feedback from VLMs, which serve as a substitute for human preferences. The feedback system is designed to evaluate generated videos in terms of overall coherence, physical accuracy, and task completion. Experimental results show that AI feedback serves as the best proxy for human preferences. Similarly, VideoAgent \cite{soni2024videoagent} explores two feedback mechanisms for iteratively refining generative models to mitigate physical hallucinations. The first mechanism leverages VLM to assess video quality, guiding iterative refinement and filtering generated videos to produce action candidates for online execution. The second mechanism involves interactive feedback from real-world environments, where generated videos are translated into action control policies, executed in the physical world, and used to collect additional environmental data for further improvement. This approach introduces an offline-online feedback paradigm that enables video refinement without requiring large-scale data. In the autonomous driving interaction scene prediction, Gen-Drive\cite{huang2024gen} transforms the traditional prediction-planning paradigm into a generation-evaluation prediction paradigm, achieving feedback fine-tuning of the generative model by introducing a reward model that selects human preferences. PhyT2V\cite{xue2024phyt2v} employs LLMs to achieve a three-step iterative optimization process: 1) parsing textual prompts to extract objects and underlying physical rules; 2) back-inferencing the semantics of the generated video and evaluating its alignment with the original prompt, which serves as a reward signal for the LLM; 3) leveraging the derived rewards and physical constraints to reconstruct and refine the textual prompt. This closed-loop framework surpasses the limitations of traditional text-to-video (T2V) techniques by incorporating a self-optimization mechanism, thereby enhancing the physical plausibility and accuracy of generated content.

\textbf{Discussion.} Overall, actively cognitive-driven world simulation aims to achieve the simulation and prediction of the physical world by enabling models to interact with the environment, thereby facilitating a more comprehensive understanding of physical dynamics. This approach significantly improves generation efficiency, counterfactual prediction accuracy, and generalization ability. However, it heavily relies on large-scale datasets and lacks dedicated foundation models for physical understanding. To address these challenges, future research could overcome the dependence on large-scale datasets by incorporating more diverse training data, as suggested in \cite{kang2024farvideogenerationworld}. Additionally, enhancing the integration and synergy of multi-sensor data could enable a more comprehensive perception of the surrounding environment, as proposed in \cite{feng2024far}. Another promising direction is the development of large-scale physics foundation models to further enhance the model’s ability to understand and reason about physical phenomena.

\begin{figure*}[t]
    \centering
    \includegraphics[width=1\linewidth]{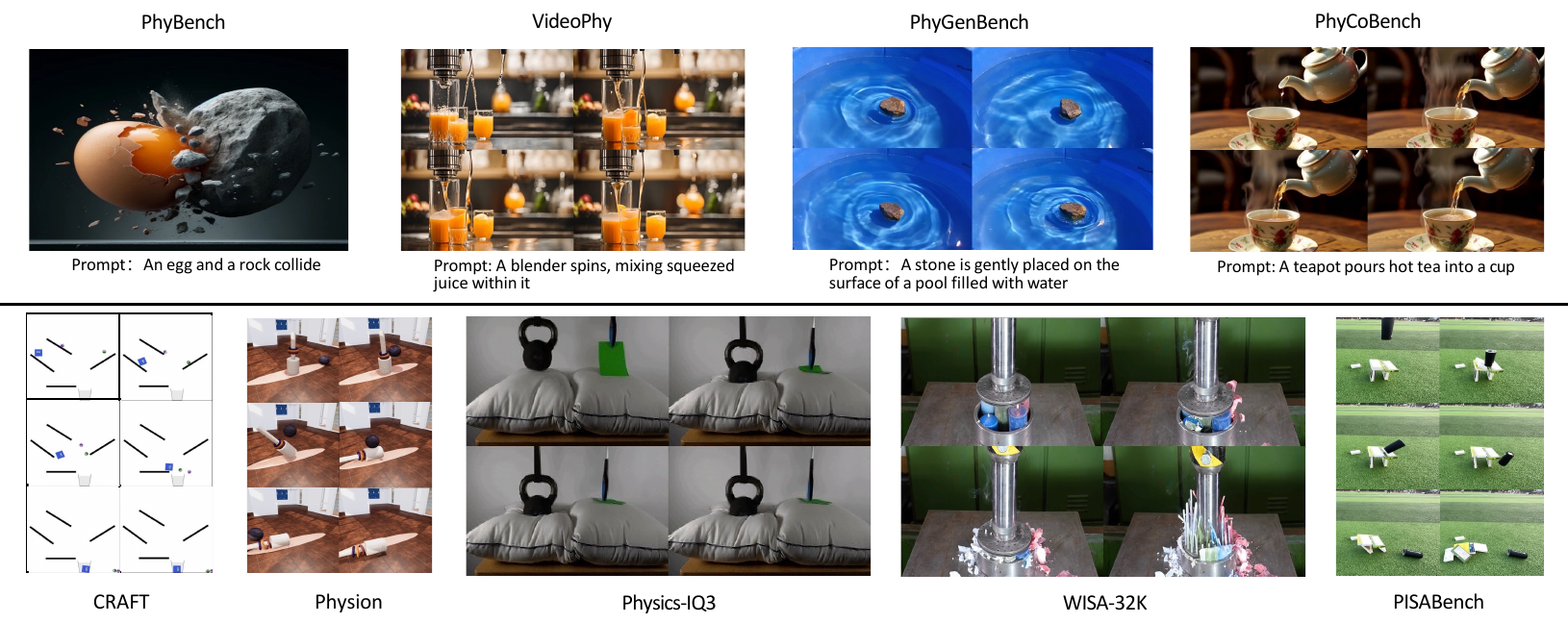}
    \caption{Physics benchmark dataset examples.}
    \label{fig:data_example}
\end{figure*}

\begin{table*}[ht]
\renewcommand{\arraystretch}{1.5}
\resizebox{\textwidth}{!}{ %
\begin{tabular}{m{0.8cm}<{\centering}m{2.5cm}<{\centering}m{1cm}<{\centering}m{1cm}<{\centering}m{1cm}<{\centering}m{2.2cm}<{\centering}m{1.5cm}<{\centering}m{1cm}<{\centering}m{1cm}<{\centering}m{1cm}}

\toprule
\multicolumn{2}{c}{\multirow{2}{*}{Dataset}} & \multicolumn{5}{c}{Physics Categories} & \multirow{2}{*}{Prompts} & \multirow{2}{*}{Videos} & \multirow{2}{*}{Scenarios} \\ 
\cline{3-7}
\noalign{\vskip -7pt}
& & Mechanics & Optics & Thermal & \makecell[c]{Material properties} & Magnetism &  &  \\ 
\noalign{\vskip -7pt}
\hline
\multirow{4}{*}{\makecell[c]{Caption\\ Level} }
& PhyBench\cite{meng2024phybench} & $\checkmark$ & $\checkmark$ & $\checkmark$ & $\checkmark$ &  & 700 & - & 31 \\
& VideoPhy\cite{bansal2024videophyevaluatingphysicalcommonsense} & $\checkmark$ &  &  & $\checkmark$ &  & 688 & - & 3 \\
& PhyGenBench\cite{meng2024towards} & $\checkmark$ & $\checkmark$ & $\checkmark$ & $\checkmark$ &  & 160 & - & 27  \\
& PhyCoBench\cite{chen2025phycobench} & $\checkmark$ &  &  & $\checkmark$ &  & 120 & - & 7  \\ 

\hline
\multirow{5}{*}{\makecell[c]{Video\\ Level}} 
& CRAFT\cite{ates2022craft} & $\checkmark$ &  &  & $\checkmark$ &  & - & 58K & 20 \\
& Physion\cite{bear2021physion} & $\checkmark$ &  &  & $\checkmark$ &  & - & 17K & 8  \\
& Physion++\cite{tung2024physion++} & $\checkmark$ &  &  & $\checkmark$ &  & - & 9.5K & 9  \\
& LLMPhy\cite{cherian2024llmphy} & $\checkmark$ &  &  & $\checkmark$ & & - & 100  \\
& Physics-IQ\cite{motamed2025generative} & $\checkmark$ & $\checkmark$ & $\checkmark$ & $\checkmark$ & $\checkmark$ & - & 396 & 66 \\ 
\hline
\multirow{2}{*}{\makecell[c]{Caption \\\& Video}}& WISA-32K\cite{wang2025wisa}&$\checkmark$&$\checkmark$&$\checkmark$&$\checkmark$&&32000&32000&17\\
& PisaBench\cite{li2025pisaexperimentsexploringphysics}&$\checkmark$&&&&&361&361&1\\
\bottomrule
\end{tabular}
}
\vspace{5pt}
\caption{Comprehensive benchmark dataset based on physical rules.}
\label{tab:compare_data}
\vspace{-12pt}
\end{table*}

\section{Benchmarks and Metrics}\label{sec:benchmark_metric}
Existing video evaluation benchmarks\cite{huang2024vbench} and metrics (e.g., PSNR\cite{hore2010image}, SSIM\cite{wang2004image}, FVD\cite{unterthiner2018towards}) primarily focus on assessing pixel-level similarity or visual and semantic quality, yet fail to effectively measure whether the video content adheres to physical laws (e.g., gravity, collision, fluid dynamics). Due to the lack of evaluation for physical plausibility, models may over-optimize visual aesthetics (e.g., texture, color, spatiotemporal consistency) while neglecting the modeling of physical principles, resulting in generated content that ``appears realistic but is physically implausible.'' Although general-purpose large-scale video datasets enable the training of versatile models, the scarcity of videos with physical properties often leads to suboptimal performance in generating physically plausible videos\cite{motamed2025generative}. In this section, we introduce the benchmarks and metrics used to evaluate the physical fidelity of video generation models.

\subsection{Benchmarks}\label{sec:benchmarks}
Existing physics benchmarks can be categorized into 
{caption-level, video-level and caption\&video level}
benchmarks. \Cref{fig:data_example} presents examples of physics benchmarks. 
{In addition,}
\cref{tab:compare_data} provides a comparative analysis of various benchmarks.

\textbf{Benchmarks for Caption Level.} Meng et al.\cite{meng2024phybench} introduced PhyBench, a comprehensive dataset designed to evaluate the understanding of physical commonsense in text-to-image (T2I) models. The dataset encompasses four major categories: mechanics, optics, thermodynamics, and material properties, comprising 31 physical scenarios and 700 textual prompts. Each scenario is enriched with fine-grained physical principles derived from textbook knowledge and GPT-4o. 
Bansal et al.\cite{bansal2024videophyevaluatingphysicalcommonsense} proposed the VideoPhy dataset, which consists of 688 textual prompts (with an average caption length of 8.5 words), covering three types of physical dynamic interactions: rigid body-rigid body, rigid body-fluid, and fluid-fluid. The dataset provides a diverse range of visual concepts and action descriptions. Compared to VideoPhy\cite{bansal2024videophyevaluatingphysicalcommonsense}, which only describes physical phenomena in text, VideoPhy-2\cite{bansal2025videophy} provides explicit annotations of physical rules. It selects 197 real-world actions related to physical commonsense(such as object interactions, sports, and physical activities) and uses LLMs to generate 3940 text prompts showcasing specific physical interactions. These prompts guide video generation, from which candidate physical rules are inferred.
PhyGenBench\cite{meng2024towards} is a benchmark specifically designed to evaluate the physical commonsense understanding of generative models. It comprises 160 textual prompts, covering 27 representative physical laws across four domains, and providing comprehensive coverage of physical phenomena. Its scope extends beyond VideoPhy\cite{bansal2024videophyevaluatingphysicalcommonsense}, which focuses solely on simple interactions between rigid bodies and fluids. 
PhyCoBench\cite{chen2025phycobench} includes seven observable physical phenomena (e.g., Newton's laws, conservation principles, collisions) and 120 corresponding benchmark evaluation prompts.

\textbf{Benchmarks for Video Level.} In addition to evaluating the physical video generation capabilities of models in text-to-video (T2V) tasks, some studies have also benchmarked their ability to understand and reason about physical dynamics. In tasks involving the prediction of conditional continuation frames (I2V/V2V), models are required to demonstrate a profound understanding of physical laws, going beyond the mere reproduction of scenes from pre-trained memory, thereby showcasing strong generalization capabilities.
CRAFT\cite{ates2022craft} is designed to evaluate models' understanding of physical forces and causal relationships between objects. It utilizes a 2D physics simulator to generate virtual scenes with 20 distinct layouts, each producing a 10-second video clip. 
Physion\cite{bear2021physion} evaluates models' comprehension of eight physical phenomena, covering rigid-soft body interactions and complex multi-component interactions. Evaluation metrics include overall accuracy, the correlation between model outputs and human responses (Pearson correlation coefficient), and Cohen's kappa index. Building upon Physion, Physion++\cite{tung2024physion++} extends the benchmark by incorporating additional mechanical properties, such as mass, friction, elasticity, and deformability, offering a more comprehensive evaluation framework.
LLMPhy\cite{cherian2024llmphy} employs a black-box optimization approach, integrating the physical knowledge of LLMs with the simulation capabilities of physics engines to form feedback loops. Additionally, the created TraySim dataset comprises 100 simulated scenarios, focusing on the task of predicting stable poses of object instances during complex interactions. The method proposed by Kang et al.\cite{kang2024farvideogenerationworld} generates videos governed by classical mechanics laws, such as uniform motion, elastic collisions, and parabolic motion, aiming to investigate whether video generation models can discover physical laws through learning from video data. It also evaluates their performance in in-distribution (ID), out-of-distribution (OOD), and compositional generalization scenarios.
The aforementioned benchmarks\cite{ates2022craft,bear2021physion,tung2024physion++,cherian2024llmphy,kang2024farvideogenerationworld} for physical video evaluation are all synthesized using physics simulators, highlighting the need for real-world videos capturing diverse and complex physical phenomena to address this limitation. The Physics-IQ\cite{motamed2025generative} dataset encompasses five domains: solid mechanics, fluid dynamics, optics, thermodynamics, and magnetism, comprising a total of 396 high-quality videos (66 scenes × 3 perspectives × 2 recordings). Each video lasts 8 seconds and covers 66 distinct physical scenarios. Physics-IQ provides the research community with comprehensive and high-quality real-world physical interactions, and is expected to significantly advance the development of video generation models in terms of physical realism.

\textbf{Benchmarks for Caption\&Video.}
The WISA-32K dataset\cite{wang2025wisa} manually collects 32,000 video samples covering three physical categories (dynamics, thermodynamics, and optics), with detailed physical annotations generated using GPT-4o mini. These annotations are categorized into textual physical descriptions, qualitative physical categories, and quantitative physical properties for subsequent model fine-tuning and training inputs. For example, a textual description such as ``A large-scale explosion generates massive smoke and dust'' has the qualitative physical categories of ``gas motion, explosion phenomena, etc.'' and the quantitative physical properties of ``Density: debris: 1 to 2.5 $g/cm^3$''.
PisaBench\cite{li2025pisaexperimentsexploringphysics} focuses on simple drop tasks to evaluate the ability of generative models to produce accurate physical phenomena. This benchmark consists of 361 real-world free-fall videos, capturing physical properties such as gravity and dynamic collisions, along with manually annotated captions. Additionally, SAM2\cite{ravi2024sam} is used to generate segmentation masks for all objects in the videos.

In summary, these datasets collectively provide comprehensive support for research on physical plausibility in video generation, spanning tasks from text-to-video generation to frame sequence prediction, thereby driving profound advancements in the field. However, nearly all benchmarks have reached a similar conclusion: current video generation models still fall short of fully capturing physical laws\cite{bansal2025videophy,kang2024farvideogenerationworld,motamed2025generative,meng2024phybench}. As a result, existing generative models remain far from becoming true world simulators.

\subsection{Metrics}\label{sec:metric}
Apart from the high-cost human evaluation, existing automated methods for assessing physical fidelity can be broadly categorized into quantitative score-based approaches\cite{chen2025phycobench,motamed2025generative,li2025pisaexperimentsexploringphysics} and automated evaluations leveraging VLMs\cite{meng2024phybench,bansal2024videocon,meng2024towards}. Quantitative metrics provide explicit numerical computing assessments of a model’s adherence to physical principles, while VLM-based evaluations enable a more flexible assessment by incorporating high-level reasoning and contextual understanding. In this section, we discuss two major paradigms of physical consistency evaluation: Quantitative Score-Based Evaluation and VLM-based Automatic Evaluation.

\textbf{Quantitative Score-Based Evaluation.} PhyCoPredictor \cite{chen2025phycobench} is a tool for automatically evaluating the physical consistency of generative models. This approach first constructs a flow-guided generative model and trains it across diverse motion scenarios. Performance evaluation is then conducted by comparing the optical flow and videos generated by the model under assessment with those produced by the reference model. Motamed et al.\cite{motamed2025generative} aim to quantify the discrepancy between generated and real videos from multiple perspectives, including spatial IoU, spatiotemporal IoU, weighted spatial IoU, and Mean Squared Error (MSE). These four metrics are integrated into a single score, the Physical IQ score, which comprehensively tracks the model’s capability in physical understanding and generation. PISA\cite{li2025pisaexperimentsexploringphysics} introduces three spatial metrics to evaluate state-of-the-art I2V models in a fundamental physical scenario--free fall. This approach assesses the accuracy of trajectories, shape fidelity, and object persistence by computing the Trajectory L2, Chamfer Distance (CD), and Intersection Over Union (IoU) between generated and real videos, respectively.

\textbf{VLM-based Automatic Evaluation.}
Meng et al.\cite{meng2024phybench} proposed PhyEvaler, an automated evaluation framework based on GPT-4o\cite{achiam2023gpt}, which generates images from input prompts and assesses them based on scene accuracy and physical correctness. This precise design of physical text prompts can also be extended to video generation tasks.
VideoCon-Physics\cite{bansal2024videocon} leverages the VIDEOCON\cite{bansal2024videocon} model and is fine-tuned using human feedback on Semantic Adherence (SA) and Physical Commonsense (PC) to achieve more accurate assessment. SA evaluates whether the video accurately depicts the entities, actions, and relationships described in the textual prompt (e.g., a red sphere rolling and colliding with a blue cube). PC assesses whether the video adheres to fundamental physical commonsense (e.g., after the collision, the sphere and the cube move in opposite directions, following the principle of momentum conservation).
The accompanying PhyGenEval\cite{meng2024towards} framework, similar to VideoCon-Physics, also evaluates from two dimensions: SA and PC. Leveraging GPT-4o, it performs key physical frame detection, physical sequence verification, and overall naturalness analysis, enabling a hierarchical assessment of physical commonsense consistency. Furthermore, based on the SA and PC scores, VideoPhy-2-Autoeval\cite{bansal2025videophy} introduces physics rule classification (evaluating whether the generated videos comply with or violate specific physical laws), thereby improving the accuracy of video generation assessment. In the construction of the automated evaluator, VideoPhy-2-Autoeval integrates human assessment knowledge to fine-tune the aforementioned VideoCon-Physics evaluation model\cite{bansal2024videocon}, enabling automated evaluation of the physical consistency of generated videos.

\section{Prospects and Challenges}\label{sec:challenge}
As video generation advances toward more realistic world simulators and world models, the development of generative systems with physical cognition capabilities presents significant potential. However, several key challenges remain to be addressed.

\textbf{Building Large Physics Foundation Models}. While current general-purpose LLMs have demonstrated cognitive breakthroughs in multi-modal understanding and reasoning, their ability to conduct in-depth research in scientific physics remains limited. Therefore, developing and evaluating dedicated large physics models (LPMs) represents a highly promising direction. Barman et al.\cite{barman2025largephysicsmodelscollaborative} provides a potential roadmap for designing physics-specific LLMs.
By integrating physics-specific knowledge into these models, the exploration of LLMs' problem-solving and innovation capabilities in the field of physics can be facilitated. Leveraging LPMs will significantly advance the development of generative models and has the potential to give rise to the next generation of world models with physical reasoning and evolutionary capabilities.

\textbf{ Advancing Physical Fidelity in World Simulators.} The development of world simulators aims to reproduce, predict, or reason about complex real-world phenomena. This requires precise modeling of physical environments, object interactions, and dynamic behaviors to ensure both physical consistency and interpretability in simulations. 
{Future works can focus on: 1) integrating LLMs or physics engines with generative models~\cite{song2024directorllm},}
allowing the system to extract and embed physical knowledge from LLMs and physics engines into the video generation process, thereby enhancing physical realism. 
{2)
incorporating reinforcement learning-based fine-tuning techniques and human feedback~\cite{yuan2023instructvideo,liu2025improving},
}
where physical rule feedback is introduced during training to dynamically correct the generative system, improving its adherence to physical laws and generalization capabilities.

\textbf{Incorporating Multi-Sensor Data. }Physical perception extends beyond vision and text, encompassing various forms of physical data from mechanics, thermodynamics, and material science (e.g., vibration, temperature, and tactile feedback). By integrating a wider range of sensor-derived physical data into models, this approach enables multi-level physical information embedding and allows for more comprehensive interactions with the environment. The combination of these enriched representations enhances the model's ability to understand and describe complex real-world physical phenomena.

\textbf{Data Scarcity and the Sim2Real Gap.} Although video generation systems based on physical cognition have made significant progress, they often rely on large-scale, high-quality physical scene data (either synthetic or real) to capture underlying physical patterns. At the same time, embodied agent learning (such as robotics and autonomous driving) also relies on high-precision, large-scale physical data to enable reasoning, planning, situational learning, and other AI applications. However, collecting such data is typically time-consuming and labor-intensive. Moreover, the diversity of physical phenomena necessitates a corresponding diversity in training data. Therefore, a key challenge moving forward is developing efficient methods for synthesizing large-scale, physically faithful and diverse video datasets for model training and evaluation. A promising solution is leveraging high-fidelity physics simulators, such as Cosmos\cite{agarwal2025cosmos}, to generate large-scale synthetic data, which can then be used to enhance world model training. However, it is important to acknowledge that a gap still exists between simulated environments and the real world, and bridging this gap remains an open challenge.

\textbf{Efficiency of Physical Simulation.} Physical simulators are used to solve physical states and predict target interaction states, but this approach often involves frequent numerical computations during the simulation process, resulting in significant computational overhead. This makes real-time simulation particularly challenging. 
{
Potential solutions include: 1) GPU acceleration for parallel simulation. For example, Genesis\cite{Genesis}, which integrates various physical solvers, combines GPU parallel acceleration to achieve unprecedented simulation speeds.
2) Designing efficient model layers, such as replacing the full attention layer in Transformer with a linear attention layer~\cite{wang2024lingen}. 
3) Distillation acceleration, distilling teacher model with complex arithmetic into computationally efficient student model~\cite{wang2023videolcm}.
}

{
\textbf{Physical Quality Assessment.} 
Existing methods~\cite{kong2024hunyuanvideo,zheng2024open,lin2024open} usually use some pixel/visual numerical metrics to evaluate the performance of the model, such as FID, SSIM, PSNR and FVD.
However, these numerical metrics usually cannot fully demonstrate the advantages and disadvantages of the model, and are often inconsistent with human preferences.
Therefore, recent methods have started to introduce human evaluation or benchmarks from multiple perspectives~\cite{huang2024vbench} to comprehensively and accurately evaluate the model. Nevertheless, existing methods either cannot evaluate the physical quality or require a lot of manpower. A feasible idea is to develop an automatic physical quality evaluation approach in combination with cutting-edge multimodal large language models (MLLM) to fully mine their powerful abilities of physical understanding, such as PhyEvaler\cite{meng2024phybench}. At the same time, a powerful physical evaluator can replace human preferences as a reward model to optimize the generative model (such as Videoagent\cite{soni2024videoagent}), making the generated content more consistent with physical laws.
}

\section{Conclusion}\label{sec: conclusion}
In this survey, we provide a comprehensive overview of the latest advancements in physics cognition-based video generation. We categorize existing research based on the evolutionary progression of physical cognition-ranging from schematic perception to passive and to active cognition-and offer an in-depth discussion of each category. Furthermore, we summarize the available datasets and commonly used evaluation metrics. Despite the rapid progress in this field, significant challenges remain, warranting further exploration. Looking ahead, with the advancement of AGI, physics-faithful video generation models are expected to play a crucial role as world simulators, becoming an indispensable component in the pathway toward AGI realization.

\ifCLASSOPTIONcompsoc
  \section*{Acknowledgments}
\else
  \section*{Acknowledgment}
\fi

This work is supported by the National Natural Science Foundation of China (Grant No.623B2039 and U22B2053), STI 2030-Major Projects (2022ZD0208800), NSFC General Program (Grant No. 62176215).

\ifCLASSOPTIONcaptionsoff
  \newpage
\fi

\bibliographystyle{IEEEtran}
\bibliography{bibtex/lib}

\end{document}